\documentclass[10pt,twocolumn,letterpaper]{article}

\usepackage[pagenumbers]{iccv} %
\usepackage{multirow}
\usepackage{makecell}
\usepackage{ulem}

\usepackage[most]{tcolorbox}

\newcommand{\CityGPTV}{\textit{UrbanLLaVA}}
\newcommand{\CityGPTVData}{\textit{UData}}
\newcommand{\CityGPTVTraining}{\textit{UTrain}}
\newcommand{\CityGPTVBench}{\textit{UBench}}

\newcommand{\UBenchGeoQA}{GeoQA}
\newcommand{\UBenchPrediction}{TrajPredict}
\newcommand{\UBenchNavigation}{Navigation}
\newcommand{\UBenchCouting}{SceneComp}
\newcommand{\UBenchSTVSAT}{ImgRetrieval}
\newcommand{\UBenchSATSTV}{CameraLoc}
\newcommand{\UBenchSTVAddr}{STV-Address}
\newcommand{\UBenchMark}{STV-Landmark}
\newcommand{\UBenchSATAddr}{SAT-Address}
\newcommand{\UBenchLand}{SAT-Landuse}
\newcommand{\UBenchSTVMulti}{STV-Outlier}
\newcommand{\UBenchMultiPOIs}{SceneFunc}

\definecolor{iccvblue}{rgb}{0.21,0.49,0.74}
\usepackage[pagebackref,breaklinks,colorlinks,allcolors=iccvblue]{hyperref}

\def\paperID{7300} %
\def\confName{ICCV}
\def\confYear{2025}

\title{\CityGPTV: A Multi-modal Large Language Model for Urban Intelligence with Spatial Reasoning and Understanding}

\author{
\centerline{\bf Jie Feng$^\dagger$, Shengyuan Wang$^\ddagger$, Tianhui Liu$^\S$, Yanxin Xi$^\P$, Yong Li$^\dagger$} \\
\centerline{$^\dagger$Department of Electronic Engineering, BNRist, Tsinghua University, Beijing, China} \\
\centerline{$^\ddagger$Department of Computer Science and Technology, Tsinghua University, Beijing, China} \\
\centerline{$^\S$School of Electronic and Information Engineering, Beijing Jiaotong University, China} \\
\centerline{$^\P$University of Helsinki, Finland} \\
\texttt{\centerline{\{fengjie, liyong07\}@tsinghua.edu.cn}}
}

\begin{document}
\maketitle

\begin{abstract}
Urban research involves a wide range of scenarios and tasks that require the understanding of multi-modal data. Current methods often focus on specific data types and lack a unified framework in urban field for processing them comprehensively. The recent success of multi-modal large language models (MLLMs) presents a promising opportunity to overcome this limitation. In this paper, we introduce \CityGPTV, a multi-modal large language model designed to process these four types of data simultaneously and achieve strong performance across diverse urban tasks compared with general MLLMs. In \CityGPTV, we first curate a diverse urban instruction dataset encompassing both single-modal and cross-modal urban data, spanning from location view to global view of urban environment. Additionally, we propose a multi-stage training framework that decouples spatial reasoning enhancement from domain knowledge learning, thereby improving the compatibility and downstream performance of \CityGPTV~across diverse urban tasks. Finally, we also extend existing benchmark for urban research to assess the performance of MLLMs across a wide range of urban tasks. Experimental results from three cities demonstrate that \CityGPTV~outperforms open-source and proprietary MLLMs in both single-modal tasks and complex cross-modal tasks and shows robust generalization abilities across cities. Source codes and data are openly accessible to the research community via \url{https://github.com/tsinghua-fib-lab/UrbanLLaVA}.
\end{abstract}

\section{Introduction} \label{intro}

\begin{figure}[t]
  \centering
   \includegraphics[width=1\linewidth]{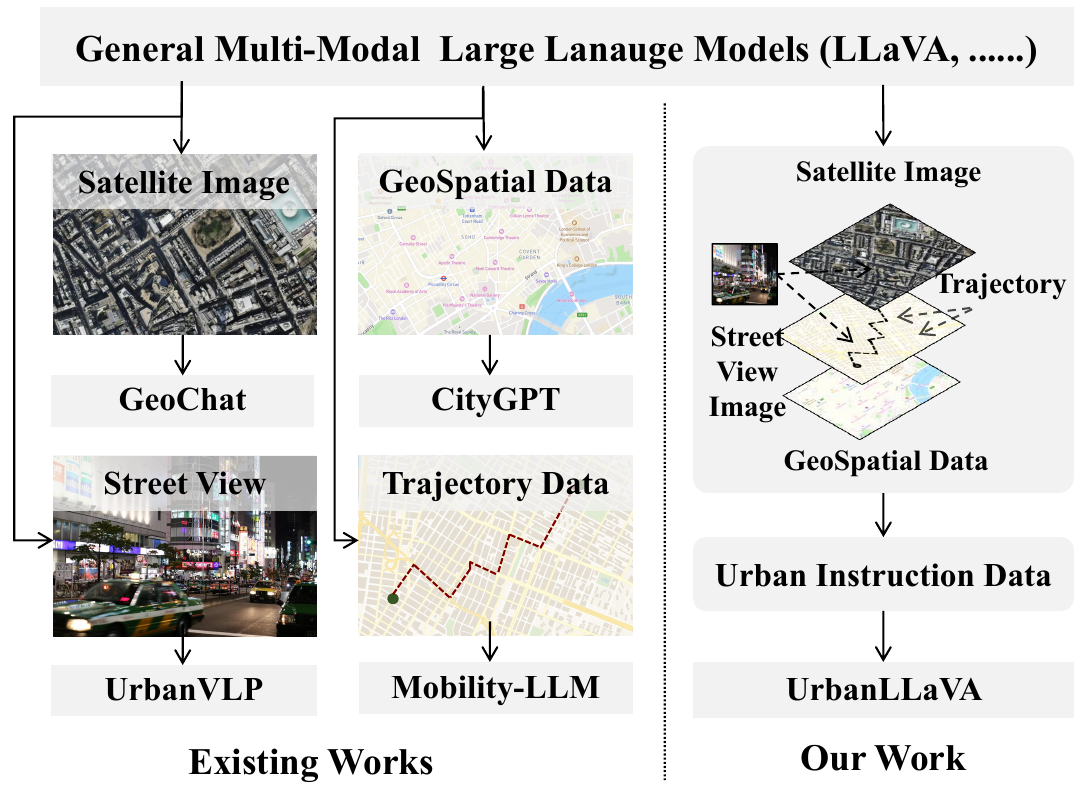}

   \caption{Existing works vs. our \CityGPTV~in urban research.}
   \label{fig:idea}
\end{figure}

Urban science~\cite{zhang2024urban, zou2025deep} and geographic science research\cite{mai2024opportunities} highlight that urban data spans multiple modalities, including urban visual data~\cite{fan2023urban}, geo-text~\cite{xiao2024analyzing}, structured geospatial data~\cite{balsebre2024city, balsebre2024lamp}, and spatiotemporal series data~\cite{gong2024mobility, li2024urbangpt}. Together, these data types capture the multi-faceted nature of urban environments, representing a wide range of spatial information and urban knowledge~\cite{mai2024opportunities, zou2025deep, xu2023urban}. Integrating these multi-modal data into a cohesive framework is essential for developing a systematic understanding of urban spaces and advancing complex modeling architectures in urban research. However, the inherent heterogeneity of these diverse urban data presents substantial challenges for the integration. While numerous deep learning based methods have been proposed to fuse various cross-domain urban data~\cite{zou2025deep}, they are often designed for specific urban tasks, limiting their ability to achieve a comprehensive understanding of urban environment and advanced reasoning for real-world urban applications~\cite{zhang2024urban, xu2023urban}.

\begin{figure*}
  \centering
   \includegraphics[width=1\linewidth]{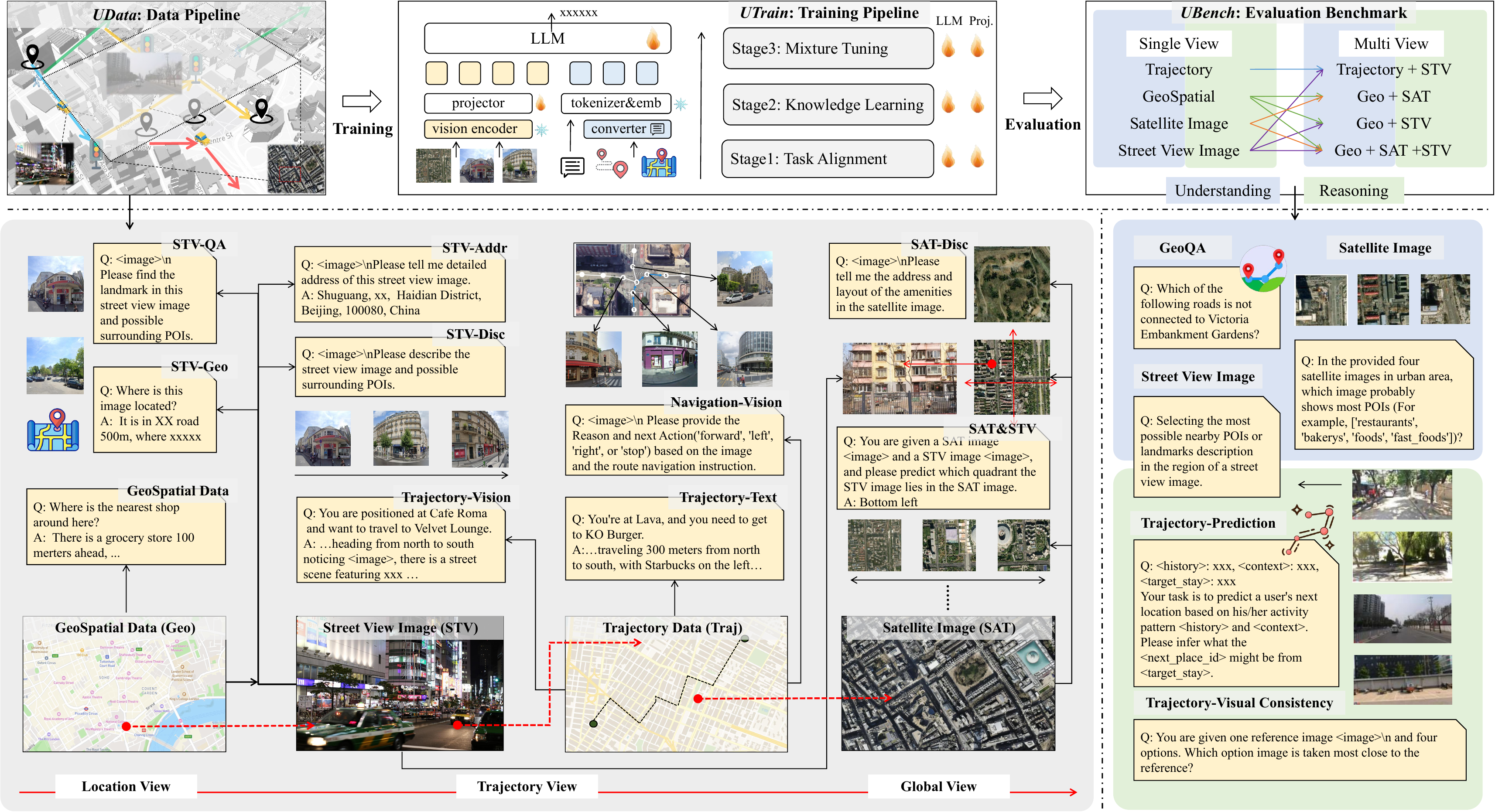}

   \caption{The framework of \CityGPTV, including \CityGPTVData, \CityGPTVTraining~and \CityGPTVBench.}
   \label{fig:framework}
\end{figure*}

Recently, multi-modal large language models (MLLMs)~\cite{yin2023survey} have made notable advancements by leveraging large language models (LLMs)~\cite{ChatGPT} with build-in common sense and reasoning abilities as a central component for unifying the processing data across various modalities, such as images~\cite{liu2024visual}, speech~\cite{girdhar2023imagebind}, and time series~\cite{jin2023time}. For example, Ma et al.~\cite{ma2023dolphins} develop a vision-language model as a conversational assistant for autonomous driving, Brohan et al.~\cite{brohan2023rt} introduce RT-2, a vision language model based end-to-end model for flexible robotics control, Li et al.~\cite{li2024llava} train LLaVA-Med for answering open-ended questions related to biomedicine images. Building on this trend, researchers have begun to explore the potential of MLLMs in urban studies~\cite{zhang2024urban}. As shown in the left part of Figure~\ref{fig:idea}, notable examples include GeoChat~\cite{kuckreja2024geochat}, an early effort in creating MLLMs for remote sensing tasks; Mobility-LLM~\cite{gong2024mobility}, which extends LLM with capabilities for trajectory modeling; and CityGPT~\cite{feng2024citygpt}, designed to process structured geospatial data with LLMs. In contrast to earlier urban data fusion methods developing in the deep learning era~\cite{zou2025deep, zhang2024urban}, these recent studies incorporate various unimodal urban data into LLMs to create obtain MLLMs that maintain the powerful reasoning abilities and address diverse urban tasks within a single modality.

However, these recent works focus solely on processing unimodal urban data and fall short of achieving a comprehensive understanding and modeling of urban system across diverse tasks involving multi-modal urban data. Unified modeling of multi-modal urban data poses significant challenges. The first challenge is the scarcity of high-quality data for cross-modality alignment. While previous works~\cite{feng2024citygpt, kuckreja2024geochat} propose various methods for constructing instruction tuning dataset for different types of unimodal urban data integrated with language, these efforts are insufficient for unified modeling across multiple modalities. A second challenge lies in the potential conflicts among diverse urban tasks across different modalities, which can lead to unstable training and inconsistent performance.

In this paper, we introduce \CityGPTV, a multi-modal large language model designed to build comprehensive urban cognition and addressing a wide range of urban tasks, which is shown in the right part of Figure~\ref{fig:idea}. In \CityGPTV, we first design \CityGPTVData, a systematic urban instruction data pipeline that enables the generation of high-quality synthetic data. In \CityGPTVData, data generation is meticulously structured to span multiple perspectives---from a localized view for single modality data to trajectory and global view for cross-modality data---capturing the inherently multi-faceted nature of urban system. To improve the training stability and model performance, we conduct extensive experiments to identify key factors impacting the training process and develop an effective three-stage training pipeline \CityGPTVTraining, based on these insights. In fact, the proposed multi-stage training framework can be viewed as a promising practice that explicitly decouples the learning of reasoning capabilities from domain-specific knowledge in MLLMs. Finally, we extend existing urban benchmarks to build a systematic urban benchmark \CityGPTVBench~for evaluating the capabilities of MLLMs in tackling diverse urban tasks. In summary, our contributions are as follows,

\begin{itemize}[leftmargin=1.5em,itemsep=0pt,parsep=0.2em,topsep=0.0em,partopsep=0.0em]
\item \CityGPTV~is the first MLLM designed for the unified modeling of four major types of urban data, with the goal of fostering comprehensive understanding and effective task-solving for urban environments, to the best of our knowledge.
\item We conduct extensive experiments to identify the key factors influencing training and propose a three-stage training pipeline that ensures stable performance of \CityGPTV~across a wide range of urban tasks involving multiple data modalities.
\item \CityGPTV~demonstrates effective integration of multi-modal data, establishing comprehensive spatial cognition and outperforming general MLLMs across various urban tasks based on results from an enhanced urban task benchmark.
\end{itemize}

\section{Methods} \label{method}

As illustrated in Figure~\ref{fig:framework}, \CityGPTV~comprises three key components: 1) the data pipeline, \CityGPTVData, designed for generating diverse and high-quality urban instruction data across various urban scenarios; 2) the training pipeline, \CityGPTVTraining, which facilitates efficient and stable training across a wide range of urban tasks; 3) the evaluation benchmark, \CityGPTVBench, for evaluating the capabilities of MLLMs in multi-modal urban tasks.

\begin{figure}[t]
  \centering
   \includegraphics[width=1\linewidth]{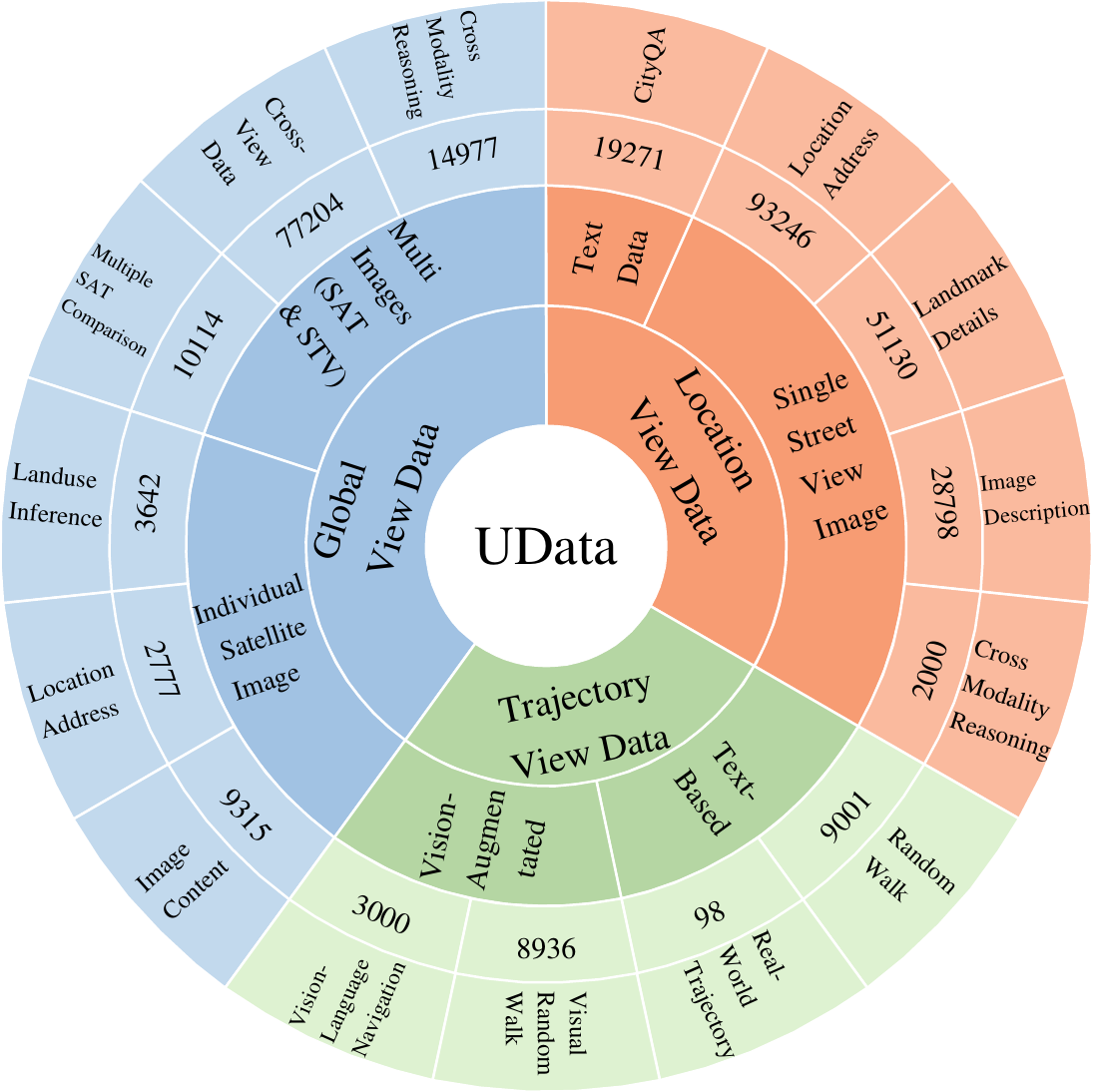}

   \caption{The thorough composition of ~\CityGPTVData~in Beijing.}
   \label{fig:umv_data}
\end{figure}

\subsection{\textbf{\CityGPTVData}: Constructing Urban Instruction Data from a Multi-View Perspective of Urban Space} \label{sec:urban-instruction}
Over the past decade, effectively integrating multi-modal urban data has emerged as a key research question in urban studies~\cite{zou2025deep}. Building on the successes of MLLMs in various fields~\cite{zhang2024urban, yin2023survey}, we extend the modelling of four types of urban data into a unified model, \CityGPTV, by constructing a diverse urban instruction data from a systematic view on the urban environment. Specifically, we organize the urban instruction data in a sequence that move from location view to a trajectory view, and finally to a global view. This approach ensures both broad spatial coverage and the integrity of relationships between different modalities in the final data. \CityGPTVData~builds upon four kinds of original urban data: 1) the structured geospatial data from OpenStreetMap~\footnote{https://www.openstreetmap.org}; 2) public trajectory data, e.g., Foursquare-checkins~\cite{yang2016participatory} and OpenStreetMap traces~\footnote{https://www.openstreetmap.org/traces}; 3) satellite images from GoogleEarth~\footnote{https://earth.google.com/}; 4) street view images from GoogleMap~\footnote{https://www.google.com/maps} and BaiduMap~\footnote{https://map.baidu.com/}. Before experiments, we collect original data from above platforms and using the following data pipeline to build instruction data. We follow the license of these platforms and ensure that the data is used only for academic research.

\subsubsection{Location View Data}
In the location view data construction stage, we focus on structured geospatial data and single street view images. Following the recent practices~\cite{balsebre2024lamp, feng2024citygpt} for structured geospatial knowledge learning, we create geospatial instruction data by designing question templates that transform basic geospatial data into natural language question and answers. For single street view image, we synthesize three types of questions: 1) two types based on predefined templates populated with information from structured geospatial data, such as, location addresses and landmark details; 2) one general MLLM generated detailed description of the image content, following the common practice for image captioning~\cite{chen2023sharegpt4v}. Throughout the data construction, we maintain a core principle of integrating street view image content with structured geographical knowledge, such as consistency in location addresses and landmark descriptions.

\subsubsection{Trajectory View Data}
Here, we construct the trajectory view data, which includes the geospatial data, trajectory data, and street view images. We start by creating two types of text-based trajectory data. The first type is generated by randomly sampling origin and destination points for routing, while the second type uses the real-world trajectory data collected from the public web source, including Foursquare-checkins and OpenStreetMap traces. To enhance geospatial context of trajectory data, we align the GPS coordinates from these original data sources with the structured geospatial data, using the textual addresses to represent locations within the trajectory. Additionally, we integrate street view images to enrich trajectory data, resulting two types of vision-augmented trajectory data. The first data extends the text-based trajectory data by incorporating street view images captured along the route (excluding intersections). We organize this data with the similar interleaved image-text format in VILA~\cite{lin2024vila}. The second data builds on the navigation instruction format akin to the classic vision-language navigation task~\cite{chen2019touchdown}. In this data, multiple street view images are presented at intersection during the trajectory, and the correct image is selected to guide the continuation of the journey.

\subsubsection{Global View Data}
Here, we present the construction of global view data designed to capture relationships among diverse data types over long distances, with street view images and satellite images as primary components and geospatial data serving as auxiliary support. Initially, we create a basic form of global view data by generating captions for single satellite image data enriched with structured geospatial data. Specifically, we define three types of data: 1) prompting general MLLM to produce detailed content description for individual satellite image; 2) sampling location address within satellite image and using a general LLM to summarize the spatial coverage of it based on these location address; 3) prompting general MLLM with land use ground-truth label to generate land use inference results with reason. 

Furthermore, we introduce the multiple satellite images for more complex instruction data. The first task is to compare the building densities across multiple satellite images. The second task focuses on identifying functional point of interest within these images. For these tasks, we provide manually crafted reasoning steps in a chain-of-thoughts format, supported by structured geospatial data, to improve the alignment between satellite images and geospatial data. Finally, we design two tasks to strengthen the alignment between the street view images and satellite images. The first task is to select the correct satellite image from a set when given a street view image, requiring the model to understand and match content or address across both image types. The second, more challenging task involves pinpointing the location of the street view image within a specific satellite image, such as identifying it as located in the top-left region of satellite image. 

Based on the data generation steps described before, we perform data quality checks and filtering on the synthesized data to ensure its quality.

\subsection{\textbf{\CityGPTVTraining}: A Multi-Stage Training Pipeline for Decoupling Reasoning and Knowledge Learning}

\begin{figure}[ht]
    \centering
    \includegraphics[width=0.48\textwidth]{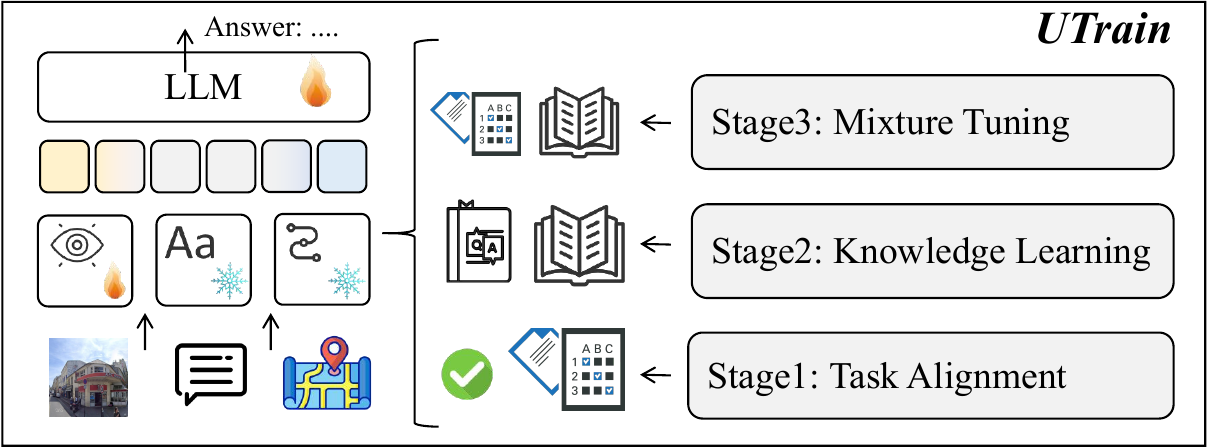}
    \caption{\CityGPTVTraining: three-stage training pipeline.}
    \label{fig:utrain}
\end{figure}

Training \CityGPTV~presents significant challenges due to the heterogeneity of multi-modal urban instruction data and the diversity of urban tasks. Achieving stable training and balancing performance across various tasks is notably difficult. We chose VILA\cite{lin2024vila} as the base model for our experiments and conduct extensive studies to identify key factors affecting training. We examine the impact of the training order of multi-modal data and trained components, observing minimal effects. However, we find that learning rate has significant effects on training stability and performance. Detailed results of them are provided in the section~\ref{sec:exp:training}. Additionally, inspired by Dong et al.~\cite{dong2023abilities}, we explore and propose an effective multi-stage training pipeline which is shown in the Figure~\ref{fig:utrain}.

We first introduce three kinds of learning procedures: \textit{knowledge learning}, \textit{task alignment} and \textit{mixture learning}. The \textit{knowledge learning} procedure refers to the training process which \CityGPTV~acquires foundational urban knowledge from various urban data, such as the information of geospatial data, pure textual trajectory, and detailed description of street view and satellite images. The \textit{task alignment} learning focuses on equipping \CityGPTV~with task-specific skills for urban applications, including vision-language navigation, trajectory prediction, chain-of-thoughts reasoning across multiple satellite and street view images. Finally, \textit{mixture learning} represents the standard training method used by most MLLMs, which training by directly mixing all types of instruction data.

During our experiments, we observe that different combination of various learning procedures significantly impact the training. Based on the observations, we propose a three-stage tuning pipeline to improve the training stability and performance on diverse urban tasks. This pipeline consists of three sequential stages:\textit{task alignment}, \textit{knowledge learning}, and finally \textit{mixture learning}. Starting with a well-trained general MLLM as our base model, we first introduce the \textit{task alignment} learning procedure, fine-tuning the model with diverse urban task related instructions to prepare it for various urban tasks. Through this phase, the model become familiar with a variety of urban tasks, leveraging its pre-existing general knowledge to complete them. However, familiarity with general knowledge alone is insufficient for effectively addressing diverse urban tasks, so we incorporate the second stage, \textit{knowledge learning} procedure. This stage imparts specialized urban knowledge from multi-modal urban data that is essential for task resolution. Finally, we introduce the mixture learning stage to enhance the model's awareness of combining knowledge and skills for solving diverse urban tasks. Here, we resample 1/3 domain specific data from the first two stages and 1/3 general textual instruction data, e.g. ShareGPT~\footnote{https://huggingface.co/datasets/shareAI/ShareGPT-Chinese-English-90k} and UltralChat~\cite{ding2023enhancing}, for final tuning.

\subsection{\textbf{\CityGPTVBench}: An Enhanced Multimodal Benchmark for Urban Intelligence Tasks}
To assess the potential of MLLMs in urban studies, CityBench~\cite{feng2024citybench} and Urbench~\cite{zhou2024urbench} have been recently introduced. Drawing from the diverse evaluation tasks in these two benchmarks, we reorganize and expand them to create the urban evaluation benchmark \CityGPTVBench, which includes 12 tasks for our experiments. All the evaluation tasks are presented in Table\ref{tab:benchmark}. We select 6 of these tasks based on the utility of the evaluation data and their relevance to urban scenarios involving \CityGPTV's urban data. For structured geospatial data and trajectory modelling, we  incorporate the GeoQA, trajectory prediction and navigation task from CityBench. For cross-view urban tasks involving both street view and satellite images, we adopt the image retrieval, camera localization, and scene comparison task from UrBench. In addition, we introduce 6 new tasks in \CityGPTVBench. Four of these tasks are designed for single street view and satellite images, including address inference for both image types, landmark recognition for street view images, and land use inference for satellite images. These single-image tasks are aligned with the urban instruction data, and we partition the original dataset into training and validation sets to prevent potential data leakage. Moreover, we build 2 additional tasks involving multiple images: 1) \UBenchSTVMulti, is a spatial consistency task for street view image, where multiple images from a single trajectory are compared to identify an outlier image not part of the trajectory; 2) \UBenchMultiPOIs, extends the scene comparison task from UrBench, challenging model to select the correct satellite image that fulfills specific functional requirements.

\begin{table}[h]
  \centering
  \setlength{\tabcolsep}{2pt} %
  \caption{Detailed information about \CityGPTVBench~for Beijing, ‘STV’ refers to street view image, and ‘SAT’ refers to satellite image.}
  \label{tab:benchmark}
  \resizebox{0.48\textwidth}{!}{
  \begin{tabular}{@{}lccccc@{}}
    \toprule
    \textbf{Tasks} & \textbf{Data} &\textbf{Category}& \textbf{Metrics} & \textbf{Samples} & \textbf{Source} \\
    \midrule
    \textbf{\UBenchGeoQA}                & Geospatial Data           & \UBenchGeoQA & Avg. Accuracy &1450 & CityBench  \\
    \textbf{\UBenchPrediction}           & Trajectory Data           & Geo+Traj & Top-1 & 500 & CityBench  \\
    \textbf{\UBenchNavigation}           & Single STV& Geo+Traj & Success Rate      & 50 & CityBench  \\
    \textbf{\UBenchCouting}       & Multi SAT& Geo+SAT & Accuracy         & 200 & UrBench    \\
    \textbf{\UBenchSTVSAT}  & Multi STV \& SAT& Geo+SS & Accuracy         & 200 & UrBench    \\
    \textbf{\UBenchSATSTV}  & Multi STV \& SAT& Geo+SS & Accuracy         & 200 & UrBench    \\
    \midrule
    \textbf{\UBenchSTVAddr}    & Single STV& Geo+STV & Accuracy         & 200 & \CityGPTVBench  \\
    \textbf{\UBenchMark} & Single STV& Geo+STV & Accuracy         & 200 & \CityGPTVBench  \\
    \textbf{\UBenchSATAddr}    & Single SAT& Geo+SAT & Accuracy         & 200 & \CityGPTVBench  \\
    \textbf{\UBenchLand}     & Single SAT & Geo+SAT & Accuracy        &  200 & \CityGPTVBench  \\
    \textbf{\UBenchSTVMulti}            & Multi STV& Geo+STV & Accuracy         & 200 & \CityGPTVBench  \\
    \textbf{\UBenchMultiPOIs}           & Multi SAT& Geo+SAT & Accuracy        & 200  & \CityGPTVBench  \\
    \bottomrule
  \end{tabular}}
\end{table}

\section{Experiments} \label{exp}
\subsection{Settings} 
We select Beijing, London and New York to conduct experiments. Due to the large volume of data, we select a region from each cities to conduct experiments. The spatial coverage of each region is shown in supplementary material. 

\textbf{MLLMs} We consider the following MLLMs as baselines: Qwen2VL-7B/72B~\cite{wang2024qwen2}, InternVL2-8B/26B~\cite{chen2024internvl, chen2024far}, VILA1.5-3B/8B/13B~\cite{lin2024vila}, LLama3.2-11B/90B~\cite{meta_llama3_2024}, and GPT4o and GPT4o-mini~\cite{openai_gpt4_2024}. For open source MLLMs, we deploy them through VLMEvalKit~\cite{duan2024vlmevalkit}. The max output tokens are set to 1000, and the temperature is set as 0.

\textbf{Metrics} Table~\ref{tab:benchmark} contains all the metrics for \CityGPTVBench. For general evaluation tasks including LLaVA-Bench(In-the-Wild)~\cite{liu2024improved}, RealWorldQA~\cite{huggingface_realworldqa_2023}, and MM-Vet~\cite{yu2024mm}, RealWorldQA uses accuracy as the metric, while LLaVA-Bench(In-the-Wild) and MM-Vet use rating score form GPT4o as the judgement.

\textbf{Implementation} We use codes from official repository~\footnote{https://github.com/NVlabs/VILA} of VILA~\cite{lin2024vila} for fine-tuning on a single 8xA100 node. The training parameters are set as follows: a learning rate of 1e-5, a maximum sequence length of 2048, a batch size of 8 per GPU, and one training epoch. Training UrbanLLaVA for Beijing on 4xA100 took a total of 10.7 hours.

\subsection{Main Results}  \label{sec:main}
\begin{table*}[h]
\centering
\setlength{\tabcolsep}{1pt} %
\footnotesize %
\renewcommand\arraystretch{1.1} %
\caption{Results on \CityGPTVBench~at Beijing, London, and New York. \CityGPTV~ significantly outperforms other baselines in most task across cities. Here, ‘STV’ denotes street view images related tasks, ‘Geo’ denotes geospatial data, ‘Traj’ denotes trajectory related task, ‘SAT’ denotes satellite images related tasks, and ‘SS’ denotes street view + satellite images. Detailed subtask and metrics can refer to Table~\ref{tab:benchmark}.}
\label{table:main_result_3cities}
\resizebox{\textwidth}{!}{
\begin{tabular}{lccccccccccccccc} 
\toprule
\textbf{City} & \multicolumn{5}{c}{\textbf{Beijing}} & \multicolumn{5}{c}{\textbf{London}} & \multicolumn{5}{c}{\textbf{New York}} \\ 
\textbf{Task Group} & \textbf{GeoQA} & \textbf{Geo+Traj} & \textbf{Geo+STV} & \textbf{Geo+SAT} & \textbf{Geo+SS} & \textbf{GeoQA} & \textbf{Geo+Traj} & \textbf{Geo+STV} & \textbf{Geo+SAT} & \textbf{Geo+SS} & \textbf{GeoQA} & \textbf{Geo+Traj} & \textbf{Geo+STV} & \textbf{Geo+SAT} & \textbf{Geo+SS} \\ 
\cmidrule(lr){2-16}
\textbf{VILA1.5-3B}    & 0.3873 & 0.0200 & 0.3967 & 0.3200 & 0.2575 & 0.4362 & 0.0400 & 0.2557 & 0.2850 & 0.2725 & 0.3954 & 0.0400 & 0.4400 & 0.2713 & 0.2425 \\
\textbf{VILA1.5-8B}    & 0.4322 & 0.0589 & 0.4300 & 0.3488 & 0.2425 & 0.4841 & 0.0884 & 0.4495 & 0.4575 & 0.2575 & 0.4575 & 0.1200 & 0.4983 & 0.3763 & 0.2525 \\
\textbf{VILA1.5-13B}   & 0.4410 & 0.1156 & 0.5167 & 0.3638 & 0.2400 & 0.4592 & 0.1298 & 0.4991 & 0.4538 & 0.2625 & 0.4501 & 0.2350 & 0.5583 & 0.4025 & 0.2825 \\
\midrule
\textbf{InternVL2-8B}  & 0.4709 & 0.1578 & 0.4667 & 0.3313 & 0.2325 & 0.4973 & 0.1347 & 0.4477 & 0.4763 & 0.2400 & 0.4632 & 0.1830 & 0.4917 & 0.4175 & 0.2400 \\
\textbf{InternVL2-26B} & 0.4877 & 0.1478 & 0.4550 & 0.3825 & 0.2275 & 0.5168 & 0.1288 & 0.4923 & 0.5138 & 0.2425 & 0.4766 & 0.2240 & 0.5217 & 0.4738 & 0.2375 \\
\midrule
\textbf{Qwen2VL-7B}    & 0.4950 & 0.1389 & 0.4383 & 0.3638 & 0.2675 & 0.4991 & 0.1560 & 0.4381 & 0.4863 & 0.2775 & 0.4567 & 0.1700 & 0.5117 & 0.5100 & 0.2950 \\
\textbf{Qwen2VL-72B}   & \uline{0.5491} & 0.1611 & \uline{0.5817} & 0.3588 & 0.2975 & 0.5802 & \uline{0.2322} & \uline{0.6375} & 0.4375 & \uline{0.3250} & 0.5273 & \uline{0.2540} & 0.6333 & 0.3788 & \uline{0.3275} \\
\midrule
\textbf{LLaMA3.2-11B}  & 0.4229 & 0.0756 & 0.4375 & 0.3075 & /      & 0.4804 & 0.1180 & 0.4000 & 0.3800 & /      & 0.4127 & 0.1100 & 0.5200 & 0.2225 & /      \\
\textbf{LLaMA3.2-90B}  & 0.4502 & 0.1056 & 0.5325 & 0.2925 & /      & 0.5659 & 0.2010 & 0.5450 & 0.4700 & /      & 0.5234 & 0.1570 & \uline{0.6825} & 0.3400 & /      \\
\midrule
\textbf{GPT4o-mini}    & 0.4542 & \uline{0.1622} & 0.4350 & 0.3800 & 0.2475 & 0.5357 & 0.1278 & 0.4752 & 0.5388 & 0.2675 & 0.5075 & 0.2320 & 0.5633 & 0.4775 & 0.2350 \\
\textbf{GPT4o}         & 0.5479 & 0.1522 & 0.4300 & \uline{0.4125} & \uline{0.3025} & \textbf{0.6446} & 0.1300 & 0.5469 & \uline{0.6050} & 0.2850 & \textbf{0.6232} & 0.2340 & 0.5767 & \uline{0.5400} & 0.2900 \\
\midrule
\textbf{\CityGPTV-VILA1.5-8B}    & \textbf{0.5682} & \textbf{0.2800} & \textbf{0.8650} & \textbf{0.6663} & \textbf{0.7025} & \uline{0.6399} & \textbf{0.2680} & \textbf{0.7500} & \textbf{0.7100} & \textbf{0.4325} & \uline{0.5773} & \textbf{0.3060} & \textbf{0.8500} & \textbf{0.7725} & \textbf{0.5825} \\
\textbf{ vs. VILA1.5-8B} & +31.47\% & +375.38\% & +101.16\% & +91.03\% & +189.69\% & +32.18\% & +203.17\% & +66.85\% & +55.19\% & +67.96\% & +26.19\% & +155.00\% & +70.57\% & +105.32\% & +130.69\% \\
\textbf{ vs. Best Baseline} & +3.48\%  & +72.63\%  & +48.70\%  & +61.53\% & +132.23\% & -0.73\%  & +15.42\%  & +17.65\% & +17.36\% & +33.08\% & -7.37\%  & +20.47\%  & +24.54\% & +43.06\% & +77.86\%  \\
\bottomrule
\end{tabular}}
\end{table*}

The main results of \CityGPTV~on three cities are presented in Table~\ref{table:main_result_3cities}, more detailed information can be accessed in \ref{sec:exp:detail_result}. We use VILA1.5-8B as the default base model in most experiments and use \CityGPTVData~with \CityGPTVTraining~methods to fine-tune it to obtain the final model \CityGPTV.

We analyze the results in Beijing first. One point to note is that, since LLama3.2 series models currently do not support multi-image input, the results for evaluation tasks involving multiple images in the \CityGPTVBench~are left blank. For models within the same series, the general trend is that larger parameter models tend to perform better, e.g., VILA1.5-13b significantly outperforms VILA1.5-3b on 5 out of 6 tasks, including both single modal and cross modal tasks. Additionally, we observe that the latest released Qwen2VL series models outperform the GPT4o series models on 2 tasks. These results demonstrate the validity and usability of our \CityGPTVBench. Our \CityGPTV~shows marked improvements over all baselines across all tasks in \CityGPTVBench. Against the best baselines, \CityGPTV~achieves performance gains ranging from 3.48\% to 132.23\% for each task. When compared to the base model VILA1.5-8B, the minimum increase is 31.47\% on the GeoQA task, while the maximum reaches an impressive 375.38\% on the Geo+Traj task. These results highlight the effectiveness of the proposed multi modal dataset, \CityGPTVData, which successfully equips smaller MLLMs with a variety of capabilities within urban space, achieving superior performance over all advanced general MLLMs.

The results in New York and London are similar to those in Beijing. Out of 5 tasks, \CityGPTV@London and \CityGPTV@NewYork both perform best in 4 tasks. However, in GeoQA task, their performance is slightly inferior to GPT4o, with reductions of -0.73\% and -7.37\%, respectively. For \CityGPTV's performance falling short of expectations on certain task, we speculate two possible reasons: first, the quality of relevant data in the two cities may be lower than that in Beijing, preventing the model from acquiring urban capabilities through learning in the training stage; second, the base model VILA1.5-8B may have comparatively weaker capabilities than commercial API GPT4o, e.g., for the GeoQA task, \CityGPTV@London outperforms VILA1.5-8B by 32.18\% but falls short of GPT4o by 0.73\%.

Overall, the proposed \CityGPTV~sucessfully enhance the performance of small MLLMs on diverse urban tasks.

\subsection{Effects of Training Strategies} \label{sec:exp:training}
Since \CityGPTV~is trained with multi-modal urban instruction data, we conduct various experiments to explore a stable and well-performing training strategies. Due to the limitation of space, we only report the multi-stage results here, more results on learning rate, modality and trained components can refer the supplementary material.

\begin{figure*}[t]
  \centering
   \begin{subfigure}[b]{0.32\textwidth}   %
        \includegraphics[width=\textwidth, height=3cm]{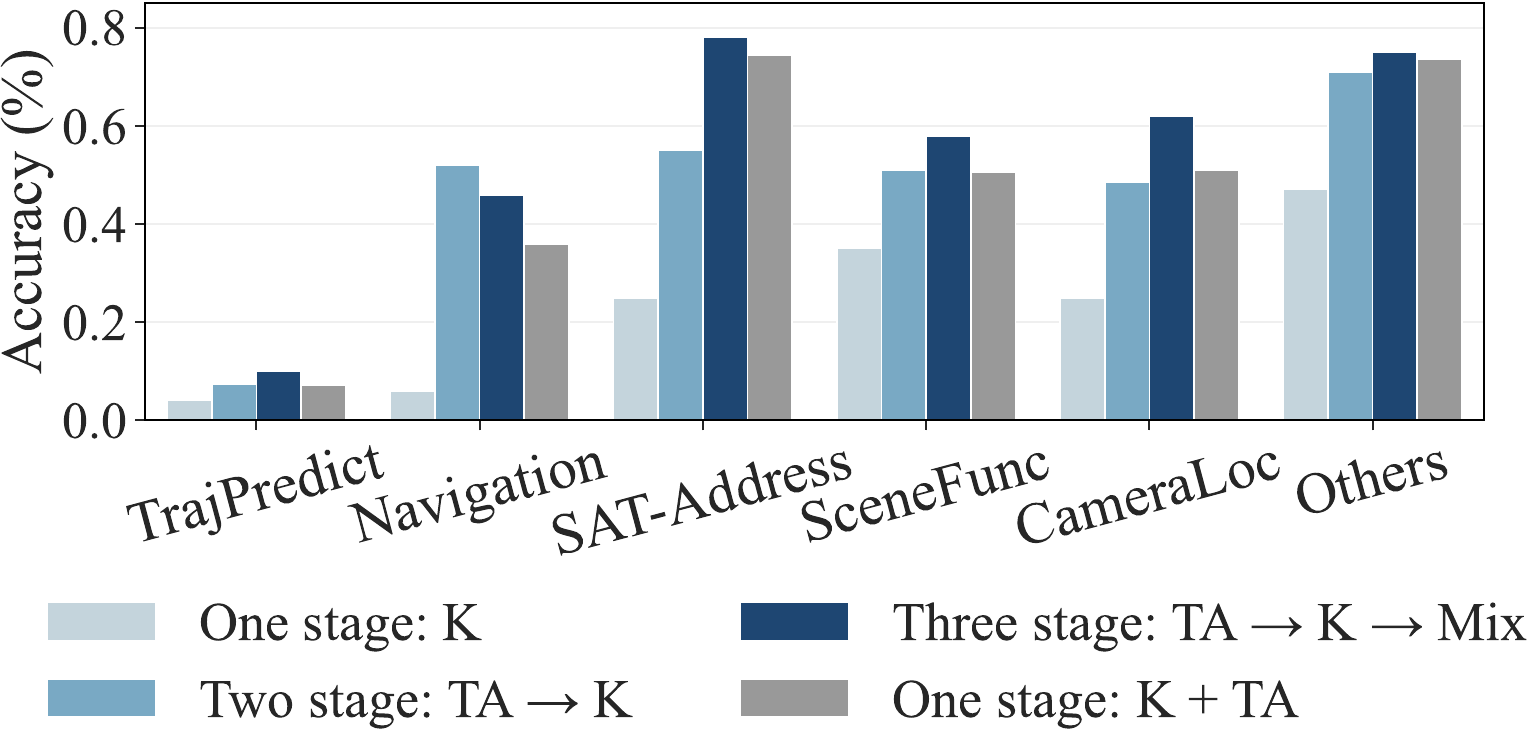} 
        \caption{The performance of three-stage tuning, gray part is the default tuning method for MLLMs.}
        \label{fig:subfig1}
    \end{subfigure}
    \hspace{1pt}  
    \begin{subfigure}[b]{0.32\textwidth}
        \includegraphics[width=\textwidth, height=3cm]{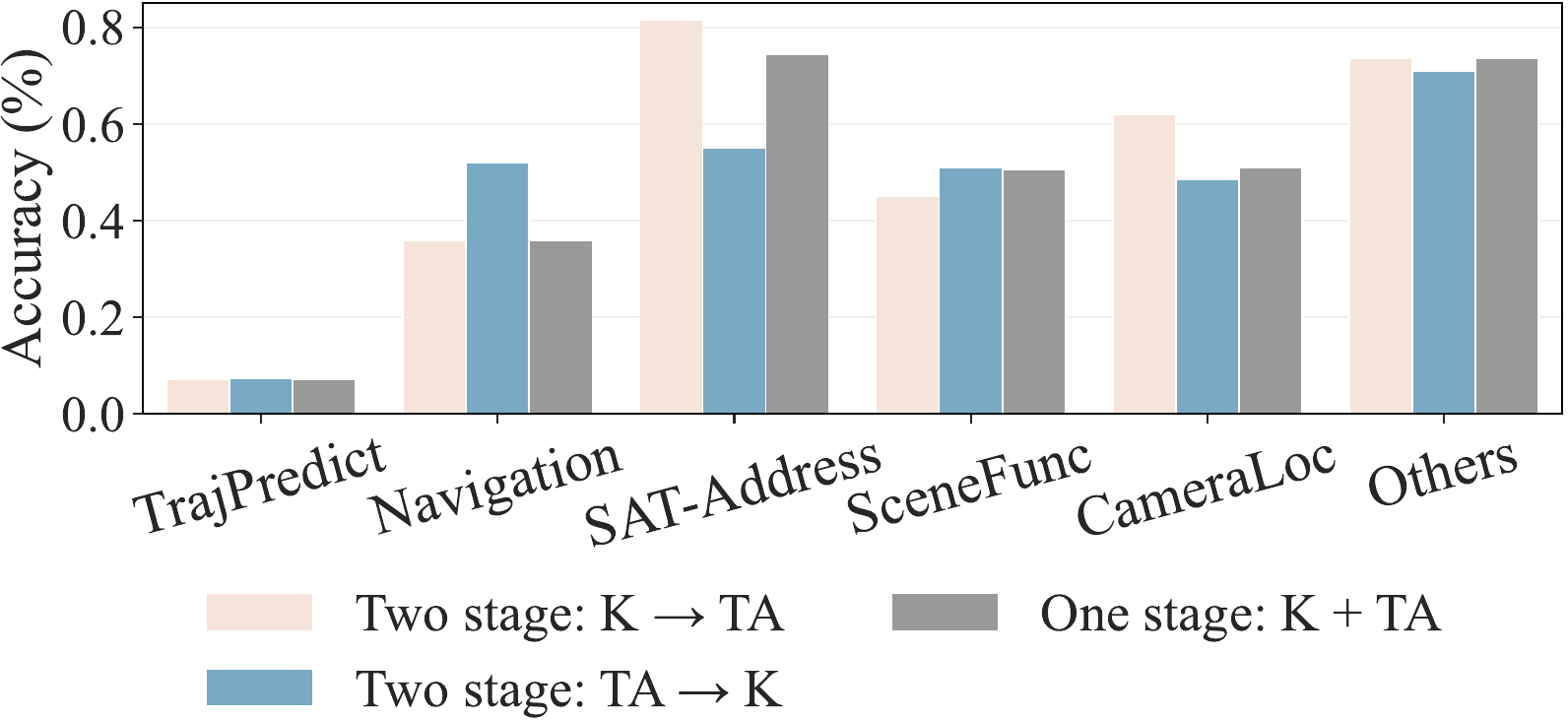} 
        \caption{The effects of the order between knowledge learning and task alignment in two-stage tuning.}
        \label{fig:subfig2}
    \end{subfigure}
    \hspace{1pt}  
    \begin{subfigure}[b]{0.32\textwidth}
        \includegraphics[width=\textwidth, height=3cm]{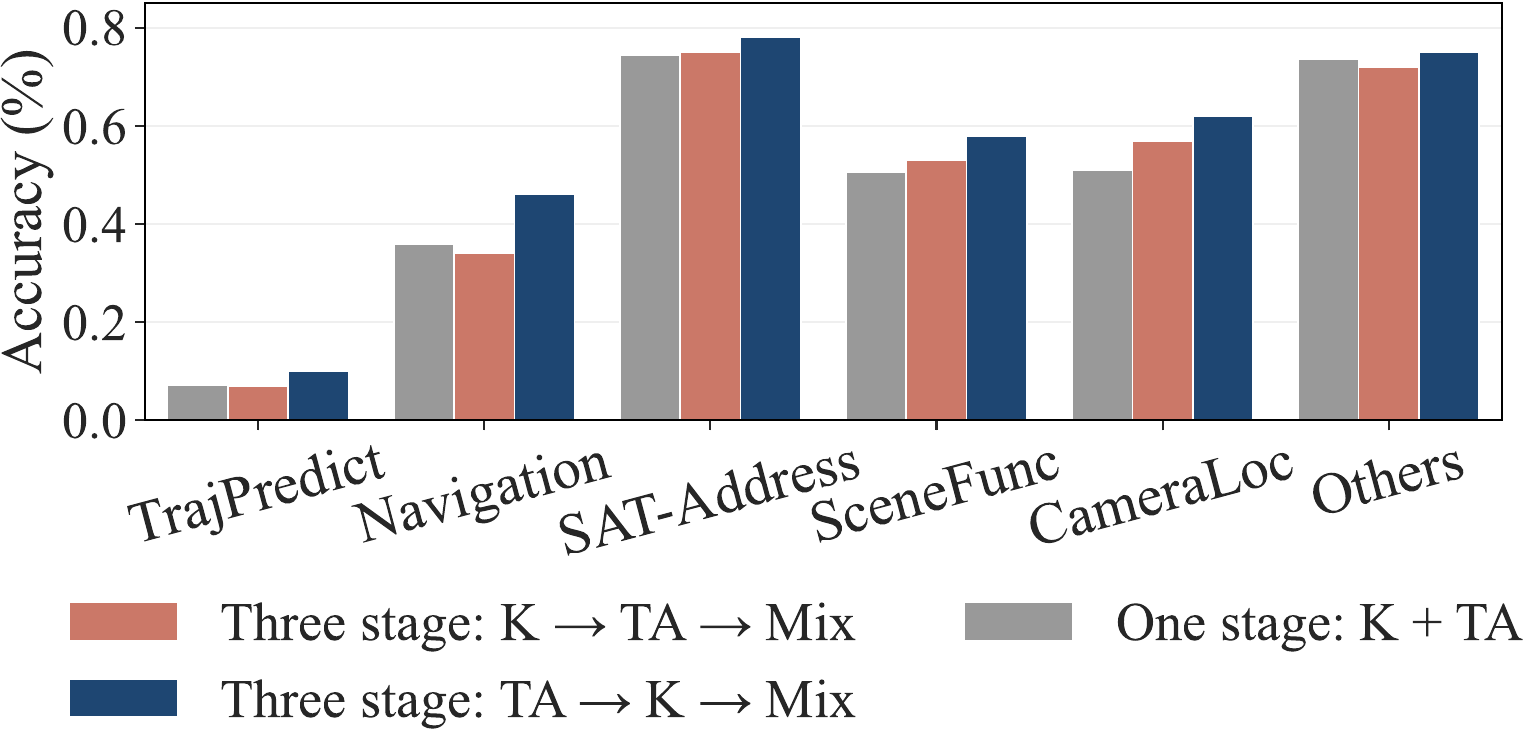} 
        \caption{The effects of the order between knowledge learning and task alignment in three-stage tuning.}
        \label{fig:subfig3}
    \end{subfigure}
    \caption{Performance of different training strategies. ‘K’ refers to \textit{knowledge learning}, ‘TA’ refers to \textit{task alignment}, and ‘Mix’ refers to \textit{mixture learning}. ‘One stage: K + TA’ means \textit{knowledge learning} and \textit{task alignment} are merged in the same stage. ‘Two stage: TA$\rightarrow$K’ means \textit{task alignment} first then \textit{knowledge learning} in the second stage. ‘Three stage: TA$\rightarrow$K$\rightarrow$Mix’ adds a step in the third stage: \textit{mixture learning}. The tasks detailed in the table are those with significant differences across different training strategies, while ‘Others’ refers to other tasks in \CityGPTVBench~with smaller differences.}
   \label{fig:stage}
\end{figure*}
We divide the training dataset into two categories: basic knowledge data and task format aligned data, aiming to develop a training pipeline that enables the model to perform stably and effectively on diverse urban tasks. As Figure~\ref{fig:subfig1} shows, ‘Three stage: TA$\rightarrow$K$\rightarrow$Mix’ performs best in most tasks and maintains reliable performance, surpassing the default tuning method for MLLMs. 
We also probe the effects of the order between \textit{knowledge learning} and \textit{task alignment} in Figure~\ref{fig:subfig2} and Figure~\ref{fig:subfig3}. We find that in two-stage training, K$\rightarrow$TA slightly outperforms TA$\rightarrow$K. However, when the third \textit{mixture learning} is added in a two-stage training model, having \textit{task alignment} first achieves better results, surpassing the models in two-stage training. We hypothesize that this is because in the two-stage training, the model first learns the foundational knowledge and then learns how to solve specific tasks. In the three-stage training, the two-stage model that \textit{knowledge learning} first and then \textit{task alignment} already possesses considerable capabilities, so the impact of mixture training is not significant. However, for two-stage model that completes \textit{task alignment} before \textit{knowledge learning}, \textit{mixture learning} can enhance its abilities, allowing it to recall how to solve urban tasks learned previously.

On the whole, the proposed three-stage training pipeline \CityGPTVTraining~integrates cross-modal data to achieve stable training and balanced performance across various urban tasks.

\subsection{Model Generalization Study} 
Here, we show that \CityGPTV~can be generalized to different data distributions and tasks, which are crucial for general urban intelligence. As Table~\ref{tab:general_benchmark_results} shows, while our \CityGPTV~performs well in diverse urban tasks, it also maintains the original stability in general scenarios, including LLaVA-Bench~\cite{liu2024improved}, RealWorldQA~\cite{huggingface_realworldqa_2023}, and MM-Vet~\cite{yu2024mm}. The results demonstrate that \CityGPTV~is competitive in the dimension of various daily-life visual tasks, real-world spatial understanding and integrated capabilities which is the base for general urban intelligence.

\begin{table}[h]
  \centering
  \caption{General benchmark results. Rating Score refers to result from the LLM-as-a-judge method with GPT4o. For LLaVA-Bench, scores range from 0 to 100, for MM-Vet, scores range from 0.0 to 1.0. Higher scores indicate better performance.
  }  
  \resizebox{0.45\textwidth}{!}{
  \begin{tabular}{@{}lccc@{}}
    \toprule
    Test@General                  & \makecell{LLaVA-Bench \\ (In-the-Wild)} & RealWorldQA & MM-Vet \\
    \midrule
    Metric                        & Rating Score              & ACC & Rating Score   \\
    \midrule
    VILA1.5-8B                    & 60.75                     & 0.3765 & 0.3518      \\
    Ours-8B                       & 58.95                     & 0.4052 & 0.3239     \\
    \bottomrule
  \end{tabular}}

  \label{tab:general_benchmark_results}
\end{table}

\begin{figure*}[t]
  \centering
   \includegraphics[width=1\linewidth]{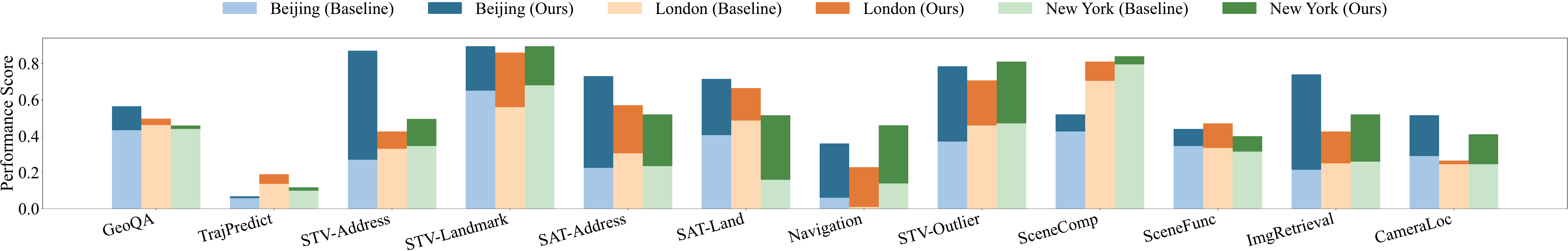}

   \caption{Learning from one city (Beijing) can be directly generalized to other cities (London and New York). In this figure, Baseline is VILA1.5-8b, and our \CityGPTV~is only trained with the urban instruction data from Beijing.}
   \label{fig:transfer_comparison}
\end{figure*}

Different cities exhibit various natural and artificial features. Thus, the transferability of urban model is important for its application.
As Figure~\ref{fig:transfer_comparison} shows, apart from in-domain capabilities empowering and performance improvement after learning, \CityGPTV~can generalize to out-of-domain tasks in various cities. Here, we examined our model trained in the Beijing training set and it exhibits competitive capabilities when tested on London and New York benchmarks. We can see from Figure~\ref{fig:transfer_comparison}, performance improvements are observed across all tasks in London and New York. Notably, for challenging aspects such as trajectory and regional tasks, the enhancements are significant, indicating the presence of similarity structures across cities that go beyond simple, naive differences. 

\subsection{Data Ablation Study}
\begin{table*}[t]
\centering
\setlength{\tabcolsep}{2pt} %
\renewcommand\arraystretch{1.1} %
\footnotesize %
\setlength{\tabcolsep}{0.6mm}
\caption{Ablation results on different urban instruction data compositions. The arrows indicate corresponding comparison with ours. Only significant differences are denoted. For \UBenchPrediction~task, the threshold is 1\%, for other tasks, the threshold is 5\%. All models are trained using the one-stage strategy to optimize experimental efficiency.}

\resizebox{\linewidth}{!}{
\begin{tabular}{@{}cccccccccccccc@{}} 
\toprule
\textbf{Task} &
  \textbf{Data View}&
  \textbf{GeoQA} &
  \textbf{TrajPredict} &
  \textbf{Address} &
  \textbf{Landmark} &
  \textbf{Address} &
  \textbf{LandUse} &
  \textbf{Navigation} &
  \textbf{STV-Outlier} &
  \textbf{SceneComp} &
  \textbf{SceneFunc} &
  \textbf{ImgRetrieval} &
  \textbf{CameraLoc} \\ \midrule
\textbf{Metric} &
  &
  \textbf{Avg. Acc} &
  \textbf{Acc@1} &
  \textbf{Acc} &
  \textbf{Acc} &
  \textbf{Acc} &
  \textbf{Acc} &
  \textbf{Success Rate} &
  \textbf{Acc} &
  \textbf{Acc} &
  \textbf{Acc} &
  \textbf{Acc} &
  \textbf{Acc} \\ \midrule
\textbf{Ours} &
    -  &
  \textbf{0.5741} &
  0.0711 &
  0.8550 &
  0.8750 &
  \textbf{0.7450} &
  \textbf{0.7850} &
  0.3600 &
  0.7800 &
  \textbf{0.5500} &
  0.5050 &
  0.7300 &
  0.5100 \\ 
\textbf{w/o CityQA} &
  Local &
  0.5409 &
  \textbf{0.0822} \textcolor{green}{$\uparrow$} &
  \textbf{0.8700}  &
  0.8900  &
  0.7150  &
  0.6950 \textcolor{red}{$\downarrow$} &
  0.4000 &
  \textbf{0.8050} &
  0.5400 &
  \textbf{0.5200} &
  0.7750 &
  0.5200 \\ 
\textbf{w/o STV} &
  Local &
  0.5192 \textcolor{red}{$\downarrow$} &
  0.0622 &
  0.4300 \textcolor{red}{$\downarrow$} &
  0.7300 \textcolor{red}{$\downarrow$} &
  0.4700 \textcolor{red}{$\downarrow$} &
  0.7200 \textcolor{red}{$\downarrow$} &
  0.4200 \textcolor{green}{$\uparrow$} &
  0.6700 \textcolor{red}{$\downarrow$} &
  0.4900 \textcolor{red}{$\downarrow$}  &
  0.4550 \textcolor{red}{$\downarrow$}  &
  0.6250 \textcolor{red}{$\downarrow$} &
  0.4250 \textcolor{red}{$\downarrow$} \\ 
\textbf{w/o Traj-Text\&Nav} &
  Trajectory &
  0.4769 \textcolor{red}{$\downarrow$}  &
  0.0644 &
  0.8100 &
  0.8800 &
  0.6350 \textcolor{red}{$\downarrow$} &
  0.7050 \textcolor{red}{$\downarrow$} &
  0.0000 \textcolor{red}{$\downarrow$} &
  0.7600 &
  0.4950 \textcolor{red}{$\downarrow$} &
  0.4300 \textcolor{red}{$\downarrow$} &
  0.6800 \textcolor{red}{$\downarrow$} &
  0.4600 \textcolor{red}{$\downarrow$} \\ 

\textbf{w/o Traj-Vision} &
  Trajectory &
  0.5590 &
  0.0690 &
  0.8350 &
  0.9050 &
  0.7300 &
  0.7100 \textcolor{red}{$\downarrow$} &
  0.3000 \textcolor{red}{$\downarrow$} &
  0.8000 &
  0.5150 &
  0.4650 &
  0.7150 &
  0.4950 \\
  
\textbf{w/o SAT-Single} &
  Global &
  0.5345 &
  0.0778 &
  0.8600 &
  \textbf{0.9100} &
  0.5550 \textcolor{red}{$\downarrow$} &
  0.4550 \textcolor{red}{$\downarrow$} &
  0.3800 &
  0.7800 &
  0.5150 &
  0.4100 \textcolor{red}{$\downarrow$} &
  0.7200 &
  0.4800 \\ 

\textbf{w/o SAT-Multi} &
  Global &
  0.5420 &
  0.0778 &
  0.8500 &
  0.8700 &
  0.6200 \textcolor{red}{$\downarrow$} &
  0.6800 \textcolor{red}{$\downarrow$} &
  0.3400 &
  0.6450 \textcolor{red}{$\downarrow$} &
  0.3500 \textcolor{red}{$\downarrow$} &
  0.3400 \textcolor{red}{$\downarrow$} &
  0.3950 \textcolor{red}{$\downarrow$} &
  0.2600 \textcolor{red}{$\downarrow$} \\   
  
  \bottomrule
\end{tabular}
}

\label{ablation_data_results}
\end{table*}

Here, we investigate the influences of different data compositions, with results shown in Table~\ref{ablation_data_results}. As outlined in Section~\ref{sec:urban-instruction}, the urban instruction data is divided into three different subsets: \textit{local view}, \textit{trajectory view}, and \textit{global view}. We remove each subset individually and observe the resulting performance differences. 
\textbf{\textit{Local view:}} It consists of textual urban geography denoted as CityQA and street view related data denoted as STV. \textit{Local view} data is important for different tasks requiring intelligence about a local part of cities. Noticeable deterioration is observed in both single-modal and multi-modal tasks, indicating the importance of locality knowledge for overall urban understanding. 
\textbf{\textit{Trajectory view:}} It describes the knowledge about continuous spaces in urban areas. It contains text-trajectory (random walk routing and real-world trajectories) and visual-trajectory( visual-language navigation instructions and random walk with visual input). 
Both text and multi-modal trajectory view datasets are essential for navigation task. It is also shown that trajectory view data is helpful for different tasks like SceneFunc and GeoQA.
\textbf{\textit{Global view:}} It includes a subset of single satellite images that focus on urban knowledge from a specific area, as well as a subset of multiple satellite images that highlight the correlations between different regions and cross alignment between satellite and street view images. Results show that they are essential to empower MLLM to handle urban tasks from a global view, e.g, ImgRetrieval and CameraLoc, while local capabilities are already competitive.

\subsection{Case Study} 
Here, we show two typical examples of urban task instances to demonstrate that \CityGPTV~can handle challenging urban tasks. Due to space limits, other cases can refer to supplementary materials.

\begin{figure}[t]
  \centering

   \includegraphics[width=1\linewidth]{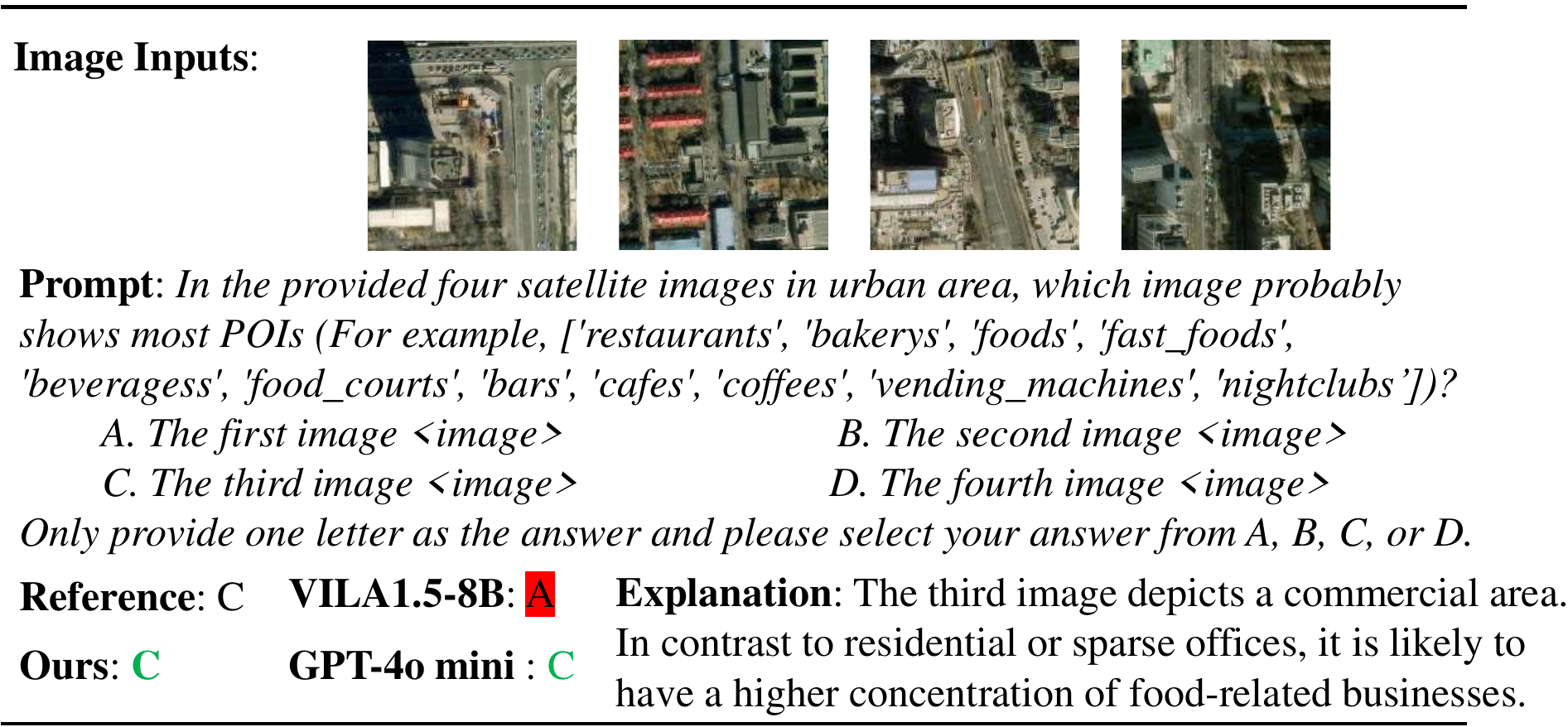}

   \caption{An example of the \UBenchMultiPOIs~task, where correct answers are in green, wrong ones in red.
   }
   \label{fig:selected_success_case_SceneFunc}
\end{figure}

\textbf{\UBenchMultiPOIs.} 
This task challenges the model to identify which satellite image contains the highest concentration of a specified category of places of interest (POIs). 
Involving multiple image inputs and text prompts, \UBenchMultiPOIs~task demands the model to understand and compare the differences between different images. As shown in Figure~\ref{fig:selected_success_case_SceneFunc}, while VILA1.5-8B fails to answer the question, our \CityGPTV~succeeds in giving the correct answer. Our model exhibits strong capabilities like multiple image understanding and comparison in this example, and is competitive with the successful closed-source model.

\begin{figure}[t]
  \centering

   \includegraphics[width=1\linewidth]{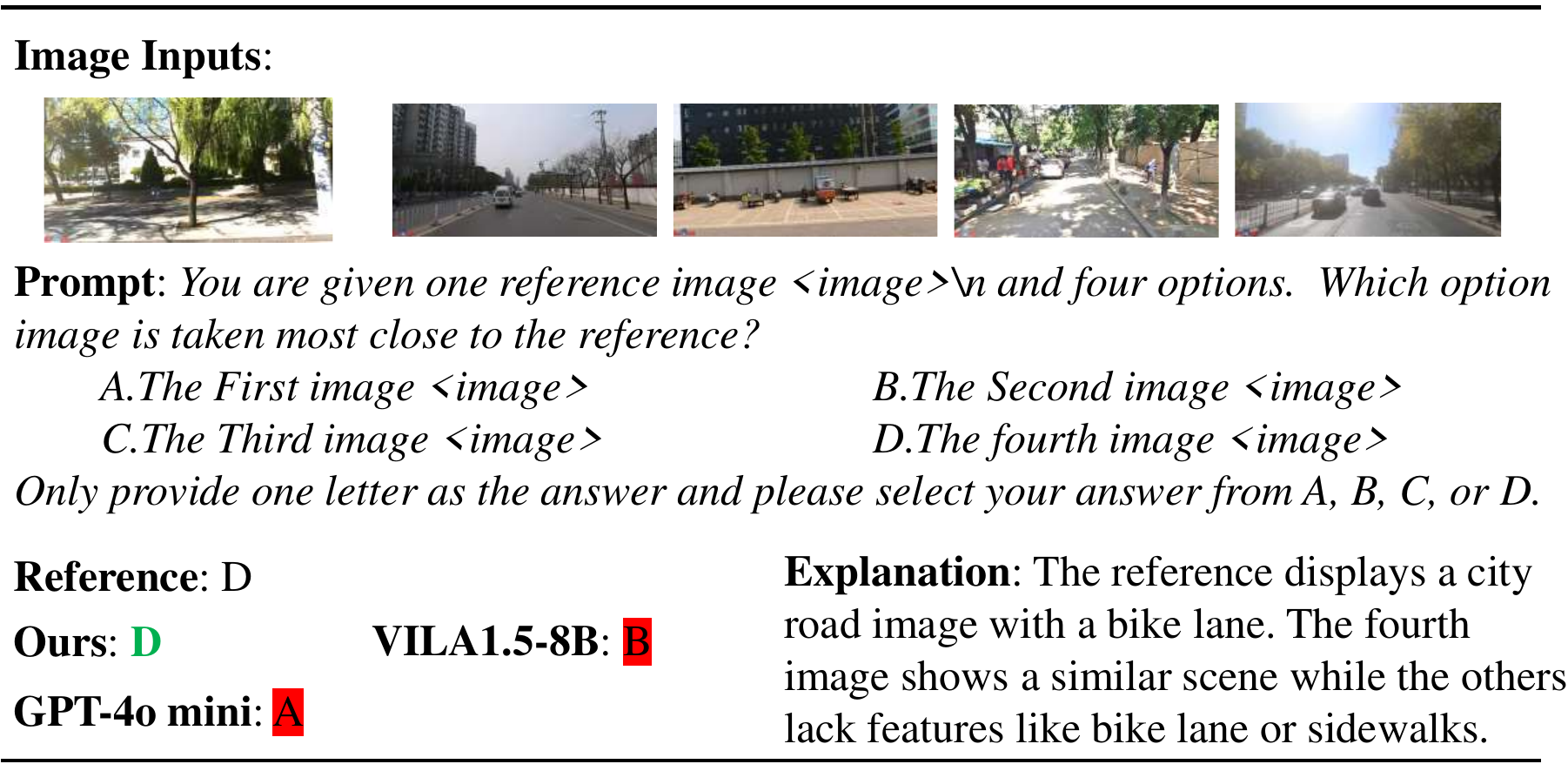}

   \caption{An example of the \UBenchSTVMulti~task.}
   \label{fig:selected_success_case_STV_Outlier}
\end{figure}

\textbf{\UBenchSTVMulti.} This task requires model to compare between multiple street views and point out the closest one to a reference. Figure~\ref{fig:selected_success_case_STV_Outlier} shows an instance of this task, where VILA1.5-8B does not successfully identify the scene of the reference image. GPT-4o-mini is closer, but it is still confused by another wrong option. Our model shows its capabilities of understanding multiple images and conducting high-level implicit logical reasoning in an urban context, outperforming these general MLLMs.

\section{Related Work} \label{relatedwork}

\subsection{Multi-modal Large Language Model}
Since the success of GPT4-V~\cite{2023GPT4VisionSC}, MLLM~\cite{yin2023survey} have become a major area of focus in research community, exemplified by the development of models like the LLaVA~\cite{liu2024visual, liu2024improved}, VILA~\cite{lin2024vila}, QwenVL~\cite{wang2024qwen2} and InternVL~\cite{chen2024internvl, chen2024far}. One of the most promising solution to develop advanced MLLM is constructing diverse and high-quality instruction dataset. 
For example, LLaVA~\cite{liu2024visual} use GPT4-V to create visual instruction tuning data, leading to the training of the first open source MLLM. Following LLaVA, VILA~\cite{lin2024vila} explore the effects of training pipelines and data formats during the pre-training stage. ShareGPT4v~\cite{chen2023sharegpt4v} further expand data scale by developing a superb caption model trained on high-quality caption data from GPT4-V. While general MLLM demonstrate strong visual understanding and reasoning capabilities~\cite{guo2024regiongpt, cheng2024spatialrgpt, wu2024v, xu2025towards} in common scenarios, they often face challenges in many specialized fields such as medical applications and remote sensing tasks. Thus, domain-specific multi-modal large language models~\cite{wu2024visionllm} are proposed, such as, Dolphins~\cite{ma2023dolphins} for autonomous driving, GeoChat~\cite{kuckreja2024geochat} for remote sensing tasks, and various models for medical application~\cite{hartsock2024visionlanguagemodelsmedicalreport}. In this paper, we propose the first MLLM for urban intelligence which can handle various data and diverse tasks in urban field.

\subsection{Multi-modal Model for Urban Study}
Urban research is an interdisciplinary field that exists multi-modal data sources~\cite{zhang2024urban, mai2024opportunities, zou2025deep, feng2025survey, ding2024understanding}, including structured geospatial data~\cite{balsebre2024city}, spatiotemporal series data~\cite{zou2025deep}, remote sensing data~\cite{mai2024opportunities, zhang2025perceiving} and street view data~\cite{biljecki2021street, fan2023urban, zhang2025urbanmllm}. Inspired by the recent advances of MLLMs, researchers explore their potential in urban studies. For structured geospatial data, Balsebre et al.~\cite{balsebre2024lamp} and Feng et al.~\cite{feng2024citygpt} propose various methods to convert structured geospatial data into a language-compatible format to enhance the geospatial knowledge of large language models. For remote sensing data~\cite{luo2024skysensegpt, muhtar2024lhrs, hu2023rsgpt, zhan2024skyeyegpt}, Kuckreja et al.~\cite{kuckreja2024geochat} and Zhang et al.~\cite{zhang2024earthgpt} design various remote sensing instruction data to fine-tune general MLLMs for various downstream remote sensing tasks. For street view data, Hao et al.~\cite{hao2024urbanvlp} fine-tune CLIP model for improved urban indicator prediction by integrating street view data and remote sensing data. Liu et al.~\cite{liu2025citylens} evaluate the potential of multi-modal language model for urban socioeconomic sensing. For spatiotemporal series data, Li et al.~\cite{li2024urbangpt} and Gong et al.~\cite{gong2024mobility} introduce domain-specific encoders to enhance LLM capabilities for spatiotemporal series modeling. Feng et al.~\cite{feng2025agentmove} propose agentic framework to unleash the power of LLM for zero-shot mobility prediction. Unlike these works that focus on limited data types and specific tasks, our method is designed to process all these data types and address a wide range of urban tasks.

\section{Conclusion}
In this paper, we propose \CityGPTV, a MLLM with enhanced urban spatial cognition by integrating four types of urban data and supporting a wide range of urban tasks. Our approach begin with the development of diverse and high-quality urban instruction data, spanning from local view to global view of urban environment. We then design a three-stage training pipeline to ensure the stable training and improved performance of model on diverse urban tasks. Experimental results from three cities on an extended urban benchmark highlight the effectiveness of \CityGPTV~for integrating multi-modal urban data and solving urban tasks. In summary, \CityGPTV~sheds lights for building the unified foundation model with powerful perception and reasoning abilities for general urban intelligence.

\section{Limitation and Future Work}
While we have made every effort to explore \CityGPTV~and present our findings clearly, several limitations remain. Our experiments have focused on the 8B model; the full potential of \CityGPTVData~and \CityGPTVTraining~on larger models has yet to be realized. In addition, \CityGPTVBench~can still be improved by refining the design of tasks, testing MLLMs' overall multi-modal capabilities from a more fine-grained perspective. Lastly, more modalities could be included like video, time series data, etc., which are also important in urban intelligence. In the future, we plan to extend \CityGPTV~to incorporate more diverse data types in urban research and tackle more advanced urban tasks from various interdisciplinary fields. 

{
    \small
    \bibliographystyle{ieeenat_fullname}
    \bibliography{references}
}

\clearpage
\setcounter{page}{1}
\maketitlesupplementary

\section{Data Example for Three Stages}
Below, we provide examples for each stage. 
\begin{figure}[ht]
    \centering
    \includegraphics[width=0.48\textwidth]{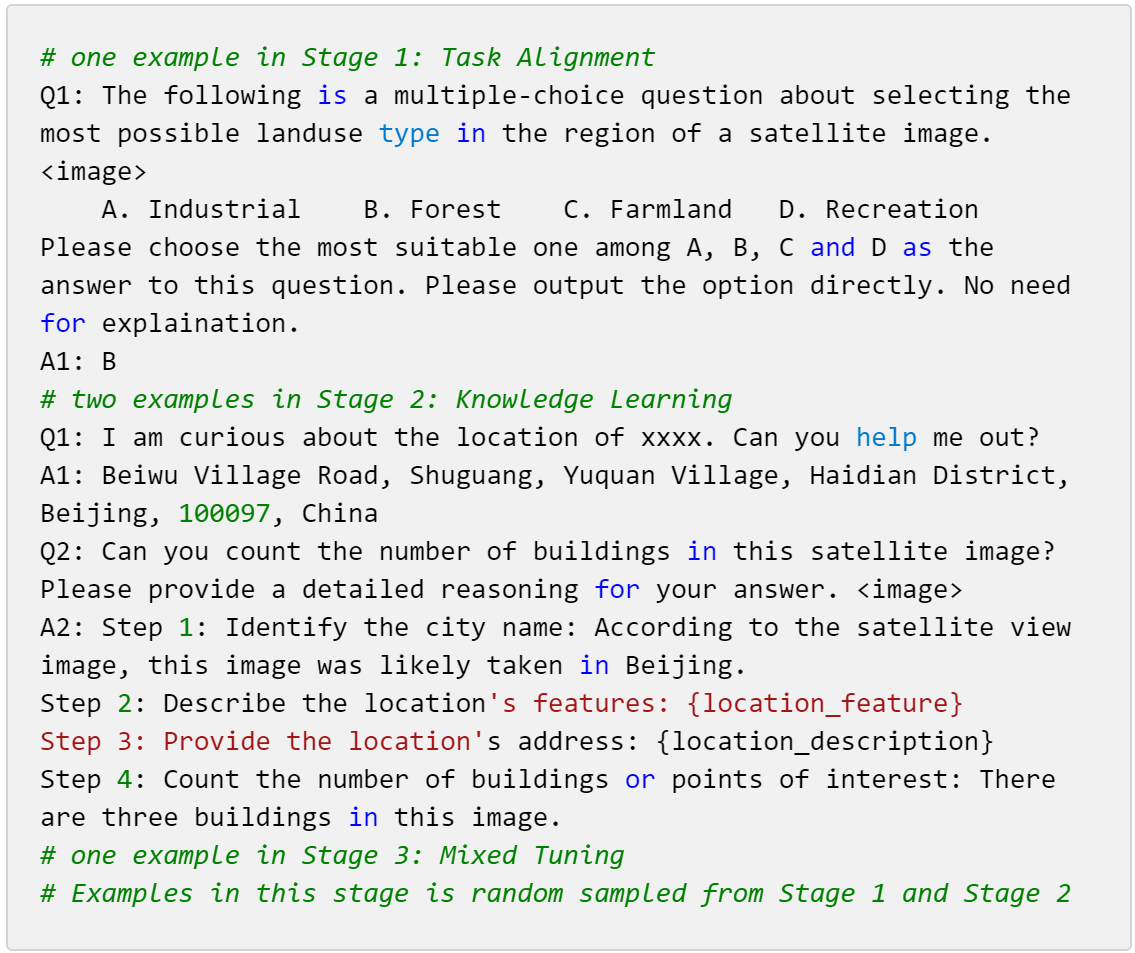}
    \caption{Input data examples for three-stage training.}
    \label{fig:input-examples}
\end{figure}

\section{Comparing with models for single-modality urban tasks}
We focus on comparing our approach with these modality-specific models to better showcase its effectiveness, which are presented in following Table~\ref{table:sinlge-compare}.
\begin{table}[ht]
\caption{Comparing with models for single-modality urban tasks.}
\centering
    \resizebox{0.49\textwidth}{!}{
    \begin{tabular}{lccccc} 
    \toprule
     & GeoQA & STV-Address & STV-Landmark & SAT-Address & SAT-Landuse \\ 
    \hline
    GeoChat & 0.3746 & 0.3100 & 0.2050 & 0.2872 & 0.4650 \\
    CityGPT & 0.5238 & - & - & - & - \\
    UrbanCLIP & - & - & - & - & 0.3750 \\
    Ours & 0.5741 & 0.8550 & 0.8750 & 0.7450 & 0.7850 \\
    \bottomrule
    \end{tabular}}
\label{table:sinlge-compare}
\end{table}

\begin{figure*}[t]
  \centering
   \begin{subfigure}[b]{0.32\textwidth}   %
        \includegraphics[width=\textwidth, height=3cm]{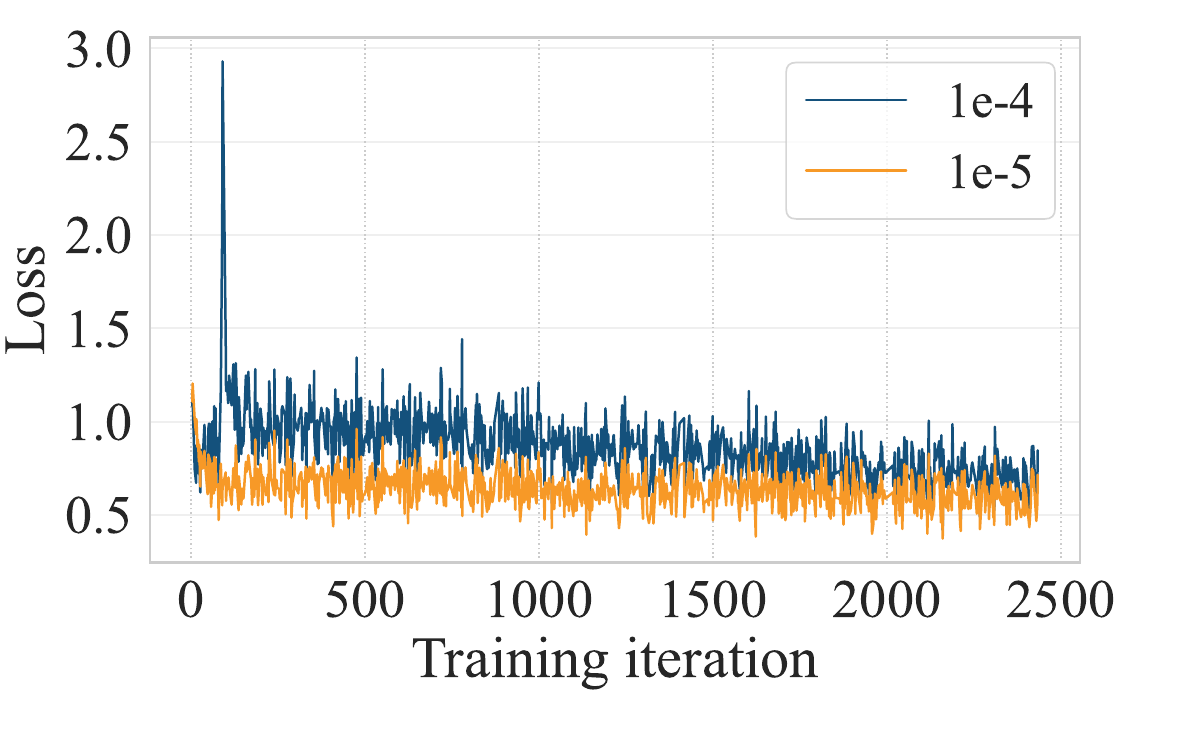} 
        \caption{The changes in training loss under the settings of learning rates $1e-4$ and $1e-5$.}
        \label{fig:learning_rate}
    \end{subfigure}
    \hspace{1pt}  
    \begin{subfigure}[b]{0.32\textwidth}
        \includegraphics[width=\textwidth, height=3cm]{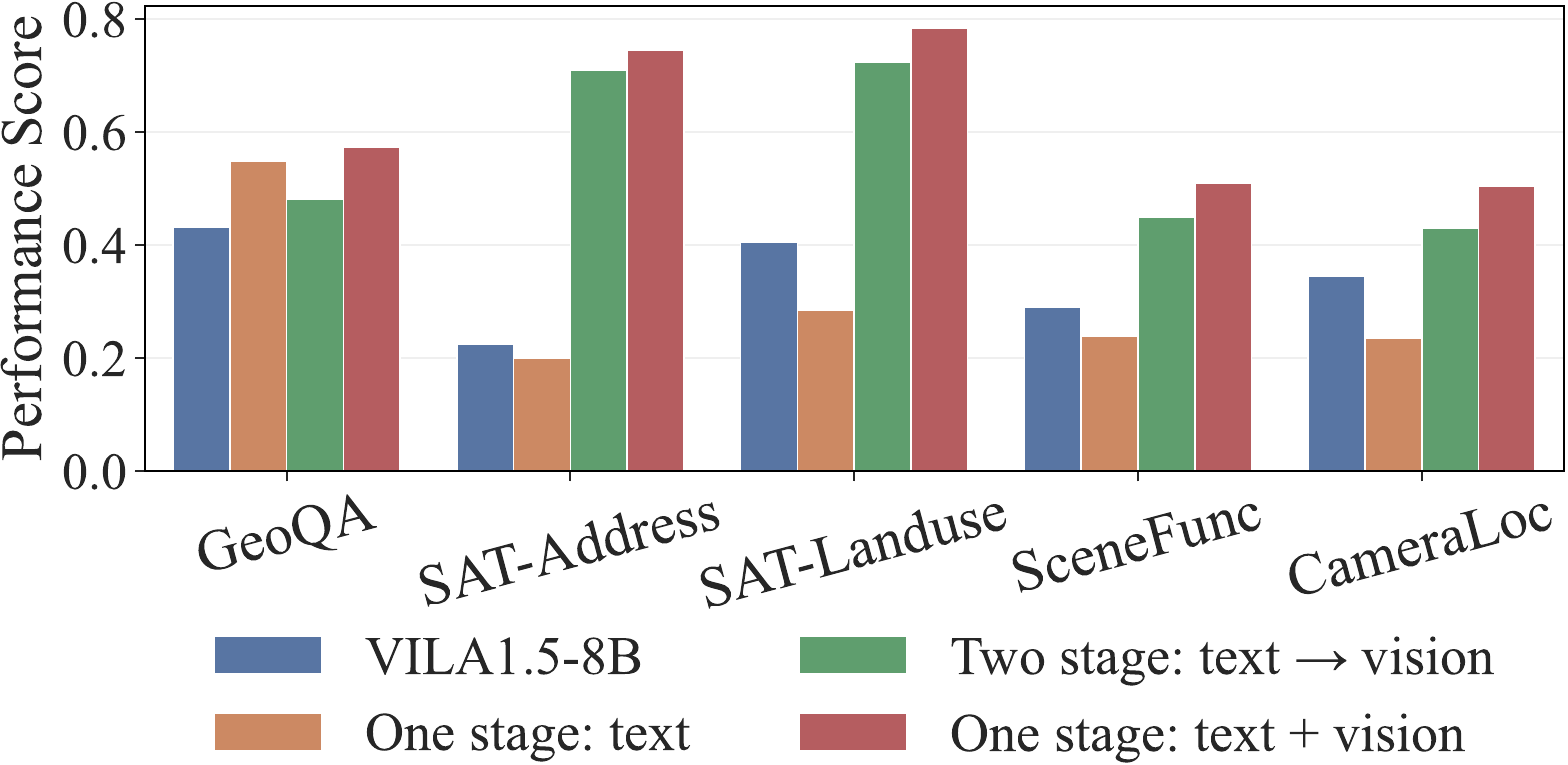} 
        \caption{Results on \CityGPTVBench~when \CityGPTVData~is divided into text data and vision data.}
        \label{fig:modality}
    \end{subfigure}
    \hspace{1pt}  
    \begin{subfigure}[b]{0.32\textwidth}
        \includegraphics[width=\textwidth, height=3cm]{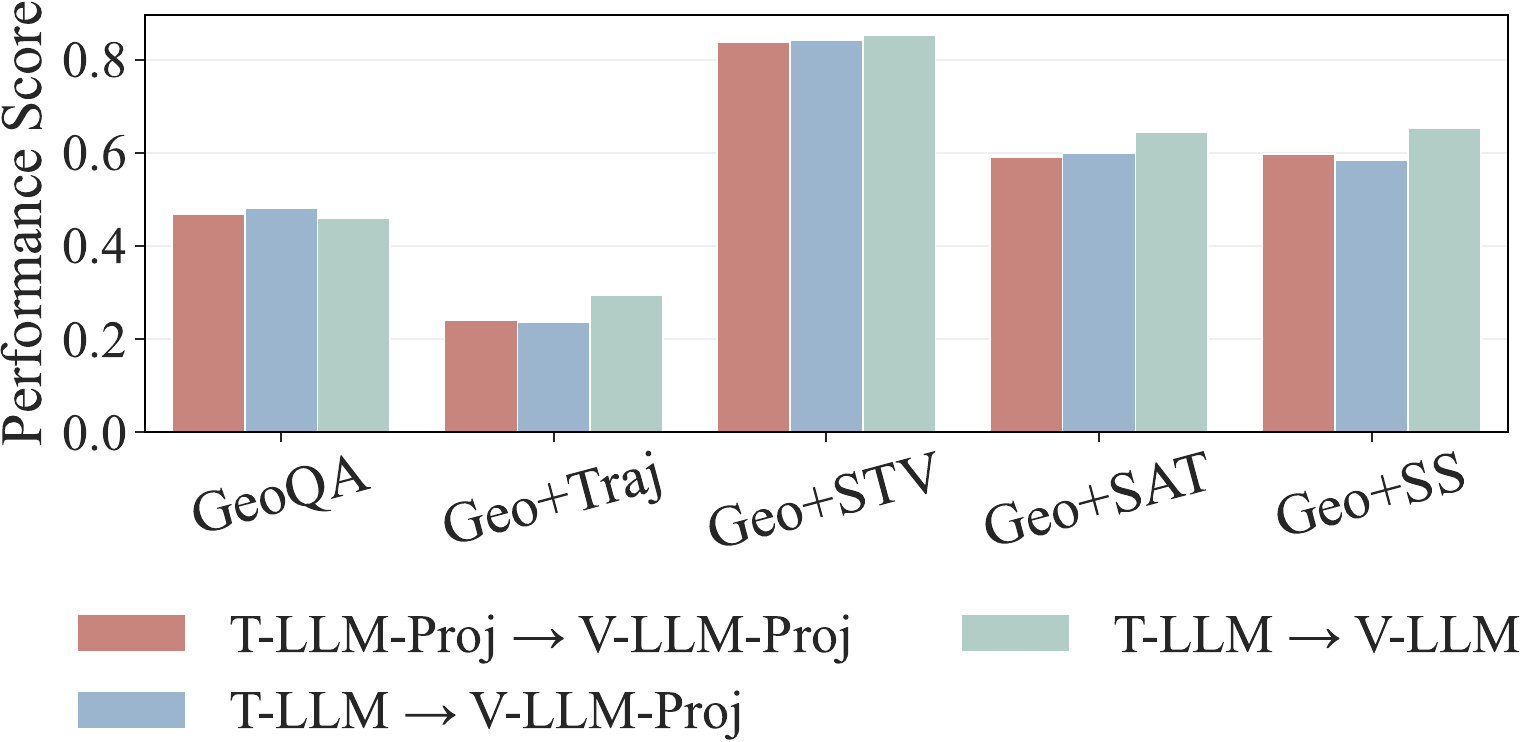} 
        \caption{Results on \CityGPTVBench~from experiments using different training components.}
        \label{fig:component}
    \end{subfigure}
    \caption{(a) illustrates that the training loss is smoother and lower when the learning rate is $1e-5$ (ours) compared to $1e-4$(VILA). (b) ‘One stage: text’ means training with text data, ‘Two stage: text$\rightarrow$vision’ means training with text data in the first stage then vision data in the second stage, ‘One stage: text+vision’ means training with text and visual data in one stage. ‘Others’ refers to other tasks in \CityGPTVBench. (c) ‘T’ refers to Text data, ‘V’ refers to Vision data and ‘T-LLM-Proj$\rightarrow$V-LLM-Proj’ means training text data with LLM and Projector in the first stage, later vision data with LLM and Projector.}
   \label{fig:stage}
\end{figure*}

\section{Additional Detailed Results of Three Cities}
\label{sec:exp:detail_result}
\begin{table*}[h]
\centering
\setlength{\tabcolsep}{2pt} %
\footnotesize %
\renewcommand\arraystretch{1.1} %
\caption{Main results on \CityGPTVBench~at Beijing. \CityGPTV~significantly outperforms other baselines in every task.}
\label{table:main_result_Bj}
\resizebox{1\textwidth}{!}{
\begin{tabular}{lccccccccccccc} 
\toprule
\multirow{2}{*}{\textbf{Tasks@Beijing}} & \textbf{GeoQA} & \multicolumn{2}{c}{\textbf{Geo+Traj}} & \multicolumn{3}{c}{\textbf{Geo+STV}} & \multicolumn{4}{c}{\textbf{Geo+SAT}} & \multicolumn{2}{c}{\textbf{Geo+SS}} \\
& \textbf{\UBenchGeoQA} & \textbf{\UBenchPrediction} & \textbf{\UBenchNavigation} & \textbf{\UBenchSTVAddr} & \textbf{\UBenchMark} & \textbf{\UBenchSTVMulti} & \textbf{\UBenchSATAddr} & \textbf{\UBenchLand} & \textbf{\UBenchCouting} & \textbf{\UBenchMultiPOIs} & \textbf{\UBenchSTVSAT} & \textbf{\UBenchSATSTV} \\
 \cmidrule(lr){2-13}
\textbf{Qwen2VL-7B} & 0.4950 & \uline{0.0978} & 0.18 & \uline{0.440} & 0.755 & 0.1200 & 0.295 & 0.405 & 0.400 & 0.355 & 0.275 & 0.260 \\
\textbf{Qwen2VL-72B} & \uline{0.5491} & 0.0822 & \uline{0.24} & 0.410 & \uline{0.785} & \uline{0.5500} & \uline{0.395} & 0.395 & 0.335 & 0.310 & 0.290 & \uline{0.305} \\
\midrule
\textbf{InternVL2-8B} & 0.4709 & 0.0957 & 0.22 & 0.420 & 0.755 & 0.2250 & 0.295 & 0.300 & 0.390 & 0.340 & 0.210 & 0.255 \\
\textbf{InternVL2-26B} & 0.4877 & 0.0756 & 0.22 & 0.440 & 0.755 & 0.1700 & 0.360 & 0.375 & 0.440 & 0.355 & 0.230 & 0.225 \\
\midrule
\textbf{VILA1.5-3B} & 0.3873 & 0.0000 & 0.04 & 0.270 & 0.655 & 0.2650 & 0.275 & \uline{0.475} & 0.295 & 0.235 & 0.250 & 0.265 \\
\textbf{VILA1.5-8B} & 0.4322 & 0.0578 & 0.06 & 0.270 & 0.650 & 0.3700 & 0.225 & 0.405 & 0.420 & 0.345 & 0.195 & 0.290 \\
\textbf{VILA1.5-13B} & 0.4410 & 0.0511 & 0.18 & 0.305 & 0.715 & 0.5300 & 0.320 & 0.320 & 0.425 & \uline{0.390} & 0.270 & 0.210 \\
\midrule
\textbf{LLaMA3.2-11B} & 0.4229 & 0.0711 & 0.08 & 0.280 & 0.595 & / & 0.290 & 0.325 & / & / & / & / \\
\textbf{LLaMA3.2-90B} & 0.4502 & 0.0711 & 0.14 & 0.295 & 0.770 & / & 0.295 & 0.290 & / & / & / & / \\
\midrule
\textbf{GPT4o-mini} & 0.4542 & 0.0844 & 0.24 & 0.280 & 0.765 & 0.2600 & 0.350 & 0.360 & \uline{0.465} & 0.345 & 0.205 & 0.290 \\
\textbf{GPT4o} & 0.5479 & 0.0844 & 0.22 & 0.405 & 0.775 & 0.1100 & 0.390 & 0.420 & 0.450 & 0.390 & \uline{0.315} & 0.290 \\
\midrule
\textbf{\CityGPTV-VILA1.5-8B} & \textbf{0.5682} & \textbf{0.1000} & \textbf{0.46} & \textbf{0.91} & \textbf{0.870} & \textbf{0.8150} & \textbf{0.780} & \textbf{0.72} & \textbf{0.585} & \textbf{0.58} & \textbf{0.785} & \textbf{0.62}  \\
\textbf{ vs. VILA1.5-8B} & +31.47\% & +73.10\% & +666.67\% & +237.04\% & +33.85\% & +120.27\% & +246.67\% & +77.78\% & +39.29\% & +68.12\% & +302.56\% & +113.79\% \\
\textbf{ vs. Best Baseline} & +3.48\% & +2.28\% & +91.67\% & +106.82\% & +10.83\% & +48.18\% & +97.47\% & +51.58\% & +25.81\% & +48.72\% & +149.21\% & +103.28\% \\ 

\bottomrule
\end{tabular}}
\end{table*}

\begin{table*}[h]
\centering
\setlength{\tabcolsep}{2pt} %
\footnotesize %
\renewcommand\arraystretch{1.1} %
\caption{Main results on \CityGPTVBench~at London. \CityGPTV~achieves better performance than other baselines in the majority of tasks.}
\label{table:main_result_Ld}
\resizebox{1\textwidth}{!}{
\begin{tabular}{lccccccccccccc} 
\toprule
\multirow{2}{*}{\textbf{Tasks@Beijing}} & \textbf{GeoQA} & \multicolumn{2}{c}{\textbf{Geo+Traj}} & \multicolumn{3}{c}{\textbf{Geo+STV}} & \multicolumn{4}{c}{\textbf{Geo+SAT}} & \multicolumn{2}{c}{\textbf{Geo+SS}} \\
& \textbf{\UBenchGeoQA} & \textbf{\UBenchPrediction} & \textbf{\UBenchNavigation} & \textbf{\UBenchSTVAddr} & \textbf{\UBenchMark} & \textbf{\UBenchSTVMulti} & \textbf{\UBenchSATAddr} & \textbf{\UBenchLand} & \textbf{\UBenchCouting} & \textbf{\UBenchMultiPOIs} & \textbf{\UBenchSTVSAT} & \textbf{\UBenchSATSTV} \\
\cmidrule{2-13}
\textbf{Qwen2VL-7B} & 0.4991 & 0.1920 & 0.12 & 0.405 & 0.760 & 0.1492 & 0.305 & 0.550 & 0.870 & 0.220 & 0.270 & \textbf{0.285} \\
\textbf{Qwen2VL-72B} & 0.5802 & \textbf{0.2245} & \uline{0.24} & 0.485 & 0.875 & \uline{0.5525} & \uline{0.530} & 0.535 & 0.420 & 0.265 & \uline{0.405} & 0.245 \\
\midrule
\textbf{InternVL2-8B} & 0.4973 & 0.1694 & 0.10 & 0.290 & 0.810 & 0.2431 & 0.315 & 0.490 & 0.785 & 0.315 & 0.215 & 0.265 \\
\textbf{InternVL2-26B} & 0.5168 & 0.1776 & 0.08 & 0.380 & 0.865 & 0.2320 & 0.355 & 0.490 &\uline{ 0.905} & 0.305 & 0.215 & \uline{0.270} \\
\midrule
\textbf{VILA1.5-3B} & 0.4362 & 0.0000 & 0.08 & 0.230 & 0.305 & 0.2320 & 0.200 & 0.445 & 0.295 & 0.200 & 0.290 & 0.255 \\
\textbf{VILA1.5-8B} & 0.4841 & 0.1367 & 0.04 & 0.330 & 0.560 & 0.4586 & 0.305 & 0.485 & 0.705 & 0.335 & 0.250 & 0.265 \\
\textbf{VILA1.5-13B} & 0.4592 & 0.1796 & 0.08 & 0.430 & 0.570 & 0.4972 & 0.275 & 0.350 & 0.800 & 0.390 & 0.275 & 0.250 \\
\midrule
\textbf{LLama3.2-11B} & 0.4804 & 0.1959 & 0.04 & 0.360 & 0.440 & / & 0.260 & 0.500 & / & / & / & / \\
\textbf{LLama3.2-90B} & 0.5659 & \uline{0.2020} & 0.20 & 0.375 & 0.715 & / & 0.385 & 0.555 & / & / & / & / \\
\midrule
\textbf{GPT4o-mini} & 0.5357 & 0.1755 & 0.08 & 0.375 & 0.835 & 0.2155 & 0.390 & 0.570 & 0.855 & 0.340 & 0.290 & 0.245 \\
\textbf{GPT4o} & \textbf{0.6446} & 0.2000 & 0.06 & \uline{0.580} & \uline{0.895} & 0.1657 & 0.480 & \uline{0.610} & 0.900 & \uline{0.430} & 0.320 & 0.250 \\
\midrule
\textbf{\CityGPTV-VILA1.5-8B} & \uline{0.6399} & 0.1959 & \textbf{0.34} & \textbf{0.610} & \textbf{0.955} & \textbf{0.6851} & \textbf{0.575} & \textbf{0.750} & \textbf{0.955} & \textbf{0.560} & \textbf{0.605} & 0.260  \\
\textbf{ vs. VILA1.5-8B} & +32.20\% & +43.28\% & +750.00\% & +84.85\% & +70.54\% & +49.40\% & +88.52\% & +54.64\% & +35.46\% & +67.16\% & +142.00\% & -1.89\% \\
\textbf{ vs. Best Baseline} & -0.72\% & -12.73\% & +41.67\% & +5.17\% & +6.70\% & +24.00\% & +8.49\% & +22.95\% & +5.52\% & +30.23\% & +49.38\% & -8.77\% \\

\bottomrule
\end{tabular}}
\end{table*}

\begin{table*}[h]
\centering
\setlength{\tabcolsep}{2pt} %
\footnotesize %
\renewcommand\arraystretch{1.1} %
\caption{Main results on \CityGPTVBench~at NewYork. \CityGPTV~achieves better performance than other models in most tasks.}
\label{table:main_result_Ny}
\resizebox{1\textwidth}{!}{
\begin{tabular}{lccccccccccccc} 
\toprule
\multirow{2}{*}{\textbf{Tasks@Beijing}} & \textbf{GeoQA} & \multicolumn{2}{c}{\textbf{Geo+Traj}} & \multicolumn{3}{c}{\textbf{Geo+STV}} & \multicolumn{4}{c}{\textbf{Geo+SAT}} & \multicolumn{2}{c}{\textbf{Geo+SS}} \\
& \textbf{\UBenchGeoQA} & \textbf{\UBenchPrediction} & \textbf{\UBenchNavigation} & \textbf{\UBenchSTVAddr} & \textbf{\UBenchMark} & \textbf{\UBenchSTVMulti} & \textbf{\UBenchSATAddr} & \textbf{\UBenchLand} & \textbf{\UBenchCouting} & \textbf{\UBenchMultiPOIs} & \textbf{\UBenchSTVSAT} & \textbf{\UBenchSATSTV} \\
\cmidrule{2-13}
\textbf{Qwen2VL-7B} & 0.4567 & 0.1200 & 0.22 & 0.585 & 0.805 & 0.1450 & 0.455 & \uline{0.395} & 0.875 & 0.315 & 0.275 & 0.315 \\
\textbf{Qwen2VL-72B} & 0.5273 & \textbf{0.1480} & \uline{0.36} & 0.550 & 0.795 & \uline{0.5550} & 0.520 & 0.235 & 0.470 & 0.290 & \uline{0.335} & \uline{0.320} \\
\midrule
\textbf{InternVL2-8B} & 0.4632 & \uline{0.1260} & 0.24 & 0.440 & 0.780 & 0.2550 & 0.395 & 0.135 & 0.835 & 0.305 & 0.245 & 0.235 \\
\textbf{InternVL2-26B} & 0.4766 & 0.1080 & 0.34 & 0.490 & 0.805 & 0.2700 & 0.495 & 0.225 & \uline{0.885} & 0.290 & 0.230 & 0.245 \\
\midrule
\textbf{VILA1.5-3B} & 0.3954 & 0.0000 & 0.08 & 0.330 & 0.745 & 0.2450 & 0.310 & 0.250 & 0.280 & 0.245 & 0.255 & 0.230 \\
\textbf{VILA1.5-8B} & 0.4575 & 0.1000 & 0.14 & 0.345 & 0.680 & 0.4700 & 0.235 & 0.160 & 0.795 & 0.315 & 0.260 & 0.245 \\
\textbf{VILA1.5-13B} & 0.4501 & 0.1100 & 0.36 & 0.375 & 0.765 & 0.5350 & 0.325 & 0.175 & 0.820 & 0.290 & 0.285 & 0.280 \\
\midrule
\textbf{LLama3.2-11B} & 0.4127 & 0.1000 & 0.12 & 0.395 & 0.645 & / & 0.295 & 0.150 & / & / & / & / \\
\textbf{LLama3.2-90B} & 0.5234 & 0.1140 & 0.20 & 0.575 & 0.790 & / & 0.460 & 0.220 & / & / & / & / \\
\midrule
\textbf{GPT4o-mini} & 0.5075 & 0.1240 & 0.34 & 0.550 & \uline{0.880} & 0.2600 & 0.415 & 0.265 & 0.880 & 0.350 & 0.255 & 0.215 \\
\textbf{GPT4o} & \textbf{0.6232} & 0.1080 & 0.36 & \uline{0.740} & 0.830 & 0.1600 & \uline{0.610} & 0.215 & \textbf{0.930} & \uline{0.405} & 0.305 & 0.275 \\
\midrule
\textbf{CityGPT-V-VILA1.5-8B} & \uline{0.5773} & 0.1120 & \textbf{0.50} & \textbf{0.920} & \textbf{0.935} & \textbf{0.6950} & \textbf{0.885} & \textbf{0.880} & 0.835 & \textbf{0.490} & \textbf{0.645} & \textbf{0.520}  \\
\textbf{ vs. VILA1.5-8B} &  +26.19\% & +12.00\% & +257.14\% & +166.67\% & +37.50\% & +47.87\% & +276.60\% & +450.00\% & +5.03\% & +55.56\% & +148.08\% & +112.24\% \\
\textbf{ vs. Best Baseline} & -7.36\% & -24.32\% & +38.89\% & +24.32\% & +6.25\% & +25.23\% & +45.08\% & +122.78\% & -10.22\% & +20.99\% & +92.54\% & +62.50\% \\ 
\bottomrule
\end{tabular}}
\end{table*}

The detailed results of \CityGPTVBench~on three cities are presented in Table~\ref{table:main_result_Bj}, Table~\ref{table:main_result_Ld} and Table~\ref{table:main_result_Ny}. Table~\ref{table:main_result_3cities} in Section~\ref{sec:main} is the aggregated results of these three tables. \CityGPTV@Beijing surpasses the baselines in all tasks, showcasing exceptional performance. \CityGPTV@London delivers top results in 9 out of 12 tasks, achieving gains over the best baseline ranging from 5.17\% to 49.38\%. Meanwhile, \CityGPTV@NewYork performs best in 9 tasks as well, with improvements over the best baseline spanning from 6.25\% to 122.78\%.

\section{Additional Results for Training Strategies}
As a supplement to results in Section~\ref{sec:exp:training}, we report results on learning rate, modality and trained components here. 

We first adjusted some experimental parameters to explore their effects, and ultimately found that the learning rate is the key parameter influencing training stability and model performance. As shown in Figure~\ref{fig:learning_rate}, we conduct experiments on the same data with different training parameters, and compared to a learning rate of 1e-4 (the default choice of VILA), the curve is smoother and lower with a learning rate of 1e-5. We think training with mixed domain-specific structured instruction data, a lower learning rate (1e-5) enables the model to handle features from different modalities more robustly.

Then we consider whether to separate text data and vision data during training to explore the influence of text data and vision data on training. But as shown in Figure ~\ref{fig:modality}, training with text and vision data in one stage yields better results compared to the other two experiments and base model VILA1.5-8B. We also investigated the impact of training components. As Figure \ref{fig:component} presents, using different components to train the same data shows little difference.

\section{Effects of Training Data Size}
\begin{figure}[hbtp]
    \centering
    \includegraphics[width=1.0\linewidth]{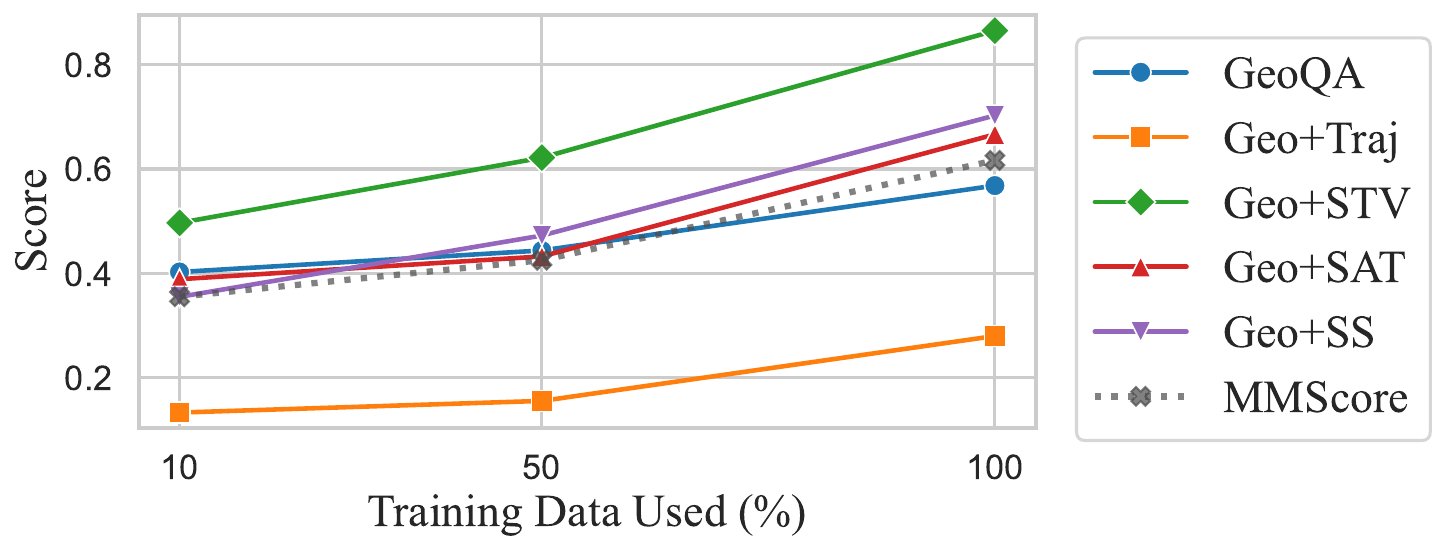}
    \caption{Scaling law from training data size to performance.}
    \label{fig:scaling}
\end{figure}

Fig. \ref{fig:scaling} presents training results with different amounts, exhibiting the high quality of UData. 

\section{Effects of Base Model}
Our method is model-agnostic and can be generalized to different MLLMs, e.g., Qwen2.5-VL-7B in Table \ref{table:qwen}.

\begin{table}[ht]
    \caption{Evaluating generalizability of methods on Qwen2.5VL.}
    \centering
    \resizebox{0.47\textwidth}{!}{
    
    \begin{tabular}{lccccc}
    \toprule
    Task Group @ Beijing                & GeoQA  & Geo+Traj & Geo+STV & Geo+SAT & Geo+SS \\ \midrule
    Qwen2.5-VL-7B-Instruct              & 0.4324 & 0.2192   & 0.4467  & 0.2850  & 0.2225 \\ 
    + Finetuned with UData             & 0.5720$\uparrow$ & 0.1876   & 0.6833$\uparrow$  & 0.4800$\uparrow$  & 0.3800$\uparrow$ \\
    \bottomrule
    \label{table:qwen}
    \end{tabular}
        }
\label{table:qwen_new}
\end{table}

\section{Effects of Model Size}
\begin{figure}[t]
  \centering

   \includegraphics[width=1\linewidth]{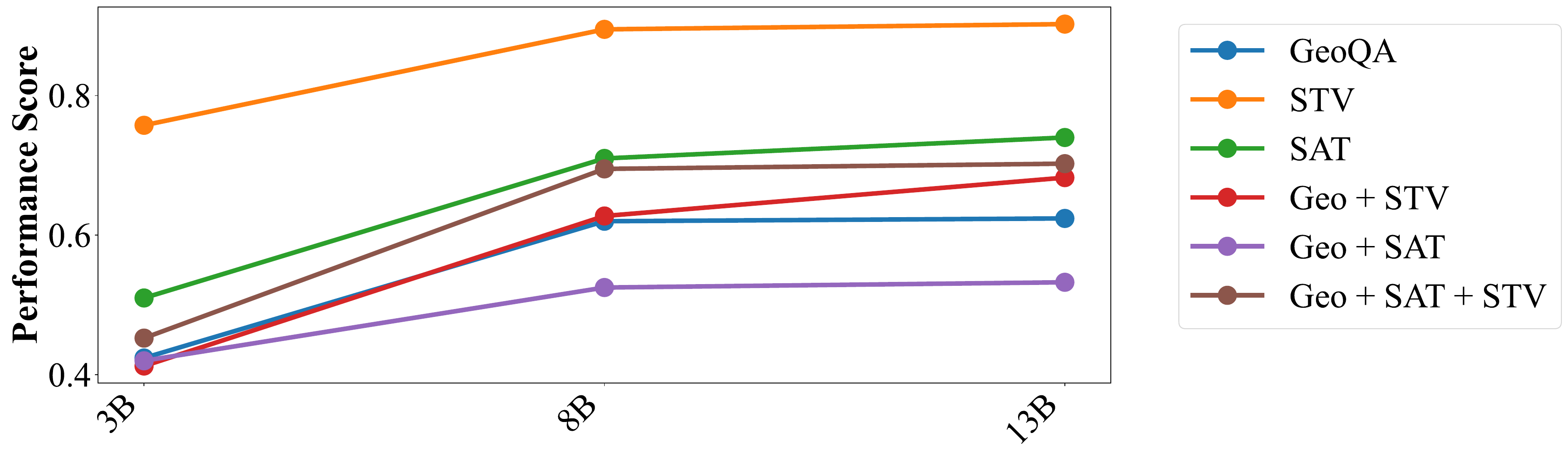}

   \caption{Results on \CityGPTV~with different model sizes.}
   \label{fig:model_performance_comparison}
\end{figure}

As Figure \ref{fig:model_performance_comparison}
shows, performance generally improves with increasing parameter size for VILA 1.5 (from 3B to 13B). However, for certain tasks, models of different sizes exhibit similar capabilities. This occurs either because the tasks are inherently challenging (e.g., trajectory prediction) or relatively easy (e.g., SAT-Landuse). Compared to VILA1.5-8B, the performance improvement of VILA1.5-13B is minimal, likely due to the capabilities of the LLaMA3-8B and LLaMA2-13B models utilized by VILA1.5. However, due to limited computing resources, we were unable to provide results for VILA1.5-40B, which could potentially achieve significantly better performance than VILA1.5-8B. 

\section{Additional Case Study}
\begin{figure}[t]
  \centering

   \includegraphics[width=1\linewidth]{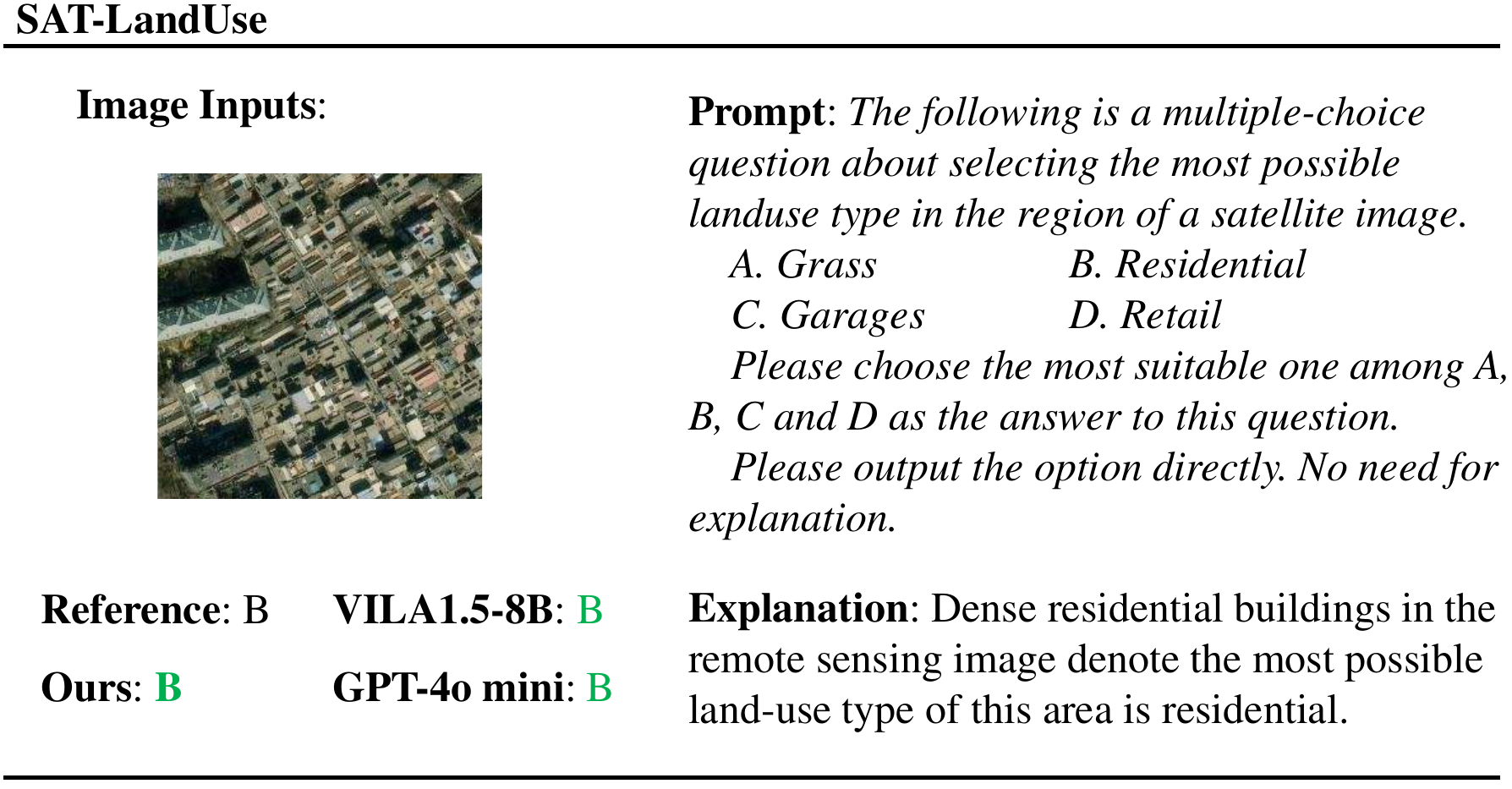}

   \caption{An example of the SAT-LandUse task. The correct answers from model are denoted with green color. The response from ours is in bold. Explanation is written by human for this question and answer.}
   \label{fig:selected_success_case_SAT_landuse}
\end{figure}

\textbf{SAT-LandUse.} This task needs a model to speculate the land use type (commercial, residential, agricultural, etc.) based on a satellite image. One example is shown in Figure \ref{fig:selected_success_case_SAT_landuse}. Our \CityGPTV~can respond to this task precisely, showing the capability of correctly perceiving the given image, satisfactory instruction following, and urban knowledge mastering. 

\begin{figure}[t]
  \centering

   \includegraphics[width=1\linewidth]{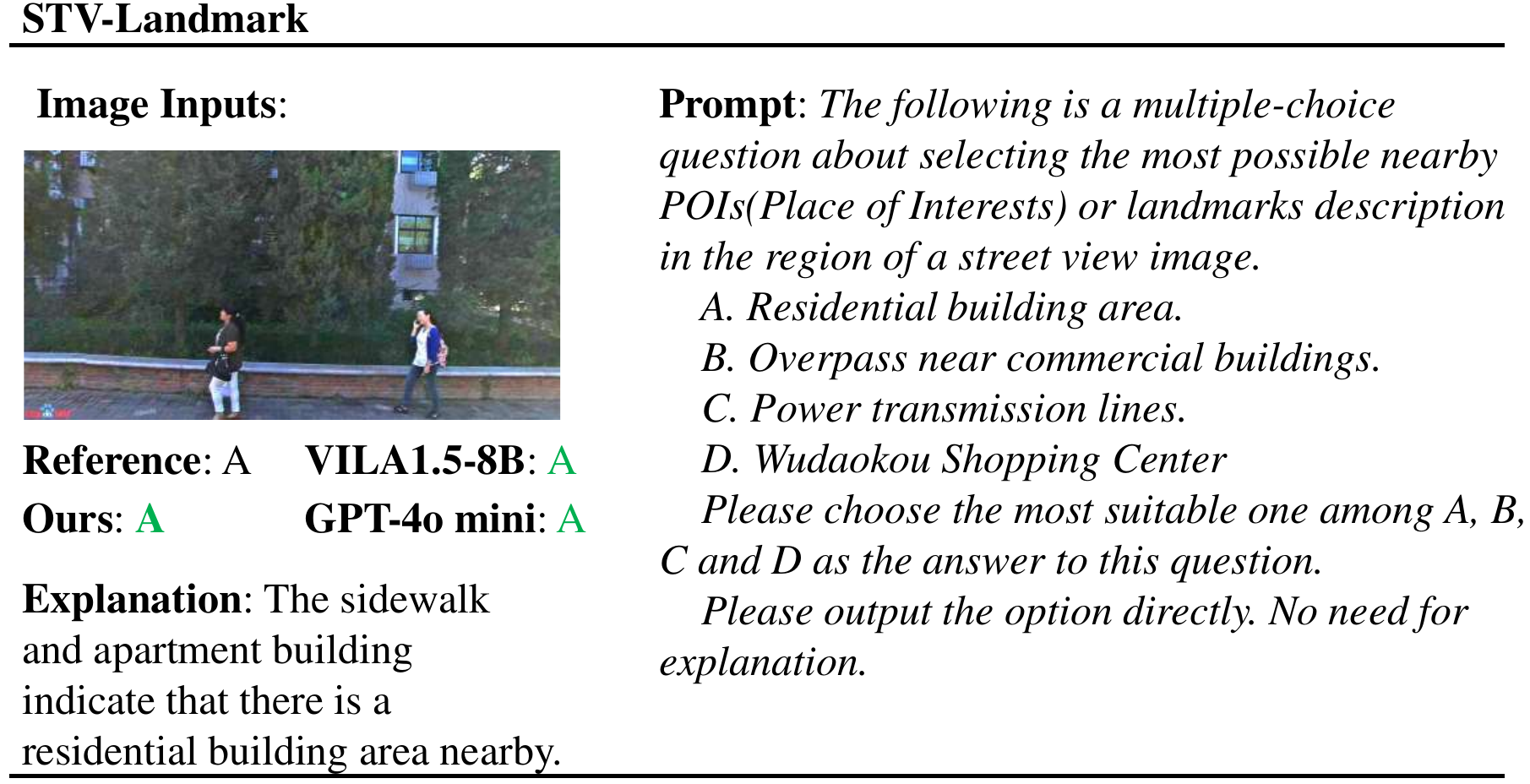}

   \caption{An example of the STV-Landmark task. The correct answers from model are denoted with green color. The response from ours is in bold. Explanation is written by human for this question and answer.}
   \label{fig:selected_success_case_STV_landmark}
\end{figure}

\textbf{STV-Landmark.} A representative example is in Figure \ref{fig:selected_success_case_STV_landmark}. In this task, models are required to find out the closest landmark feature to a given street view, which needs implicit logical reasoning capability to answer. By correctly answering a STV-Landmark question, \CityGPTV~presents its ability to conduct logical reasoning in a multi-modal context.

\begin{figure}[t]
  \centering

   \includegraphics[width=1\linewidth]{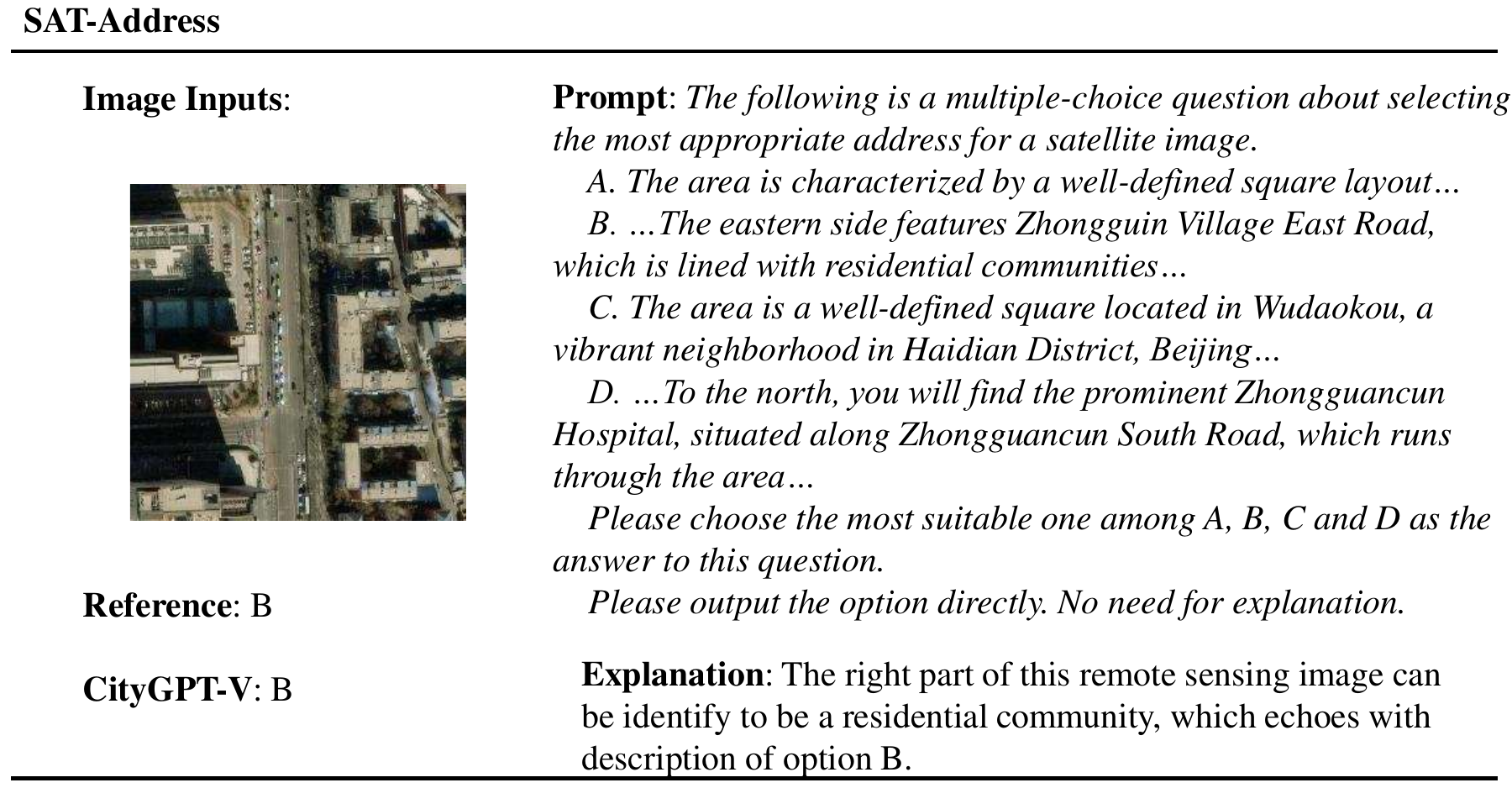}

   \caption{Example of a SAT-Address task.}
   \label{fig:success_case_SAT_address}
\end{figure}

\textbf{SAT-Address.} This task needs a model to speculate the most probable address description based on a satellite image. One example is shown in Figure \ref{fig:success_case_SAT_address}

\begin{figure}[t]
  \centering

   \includegraphics[width=1\linewidth]{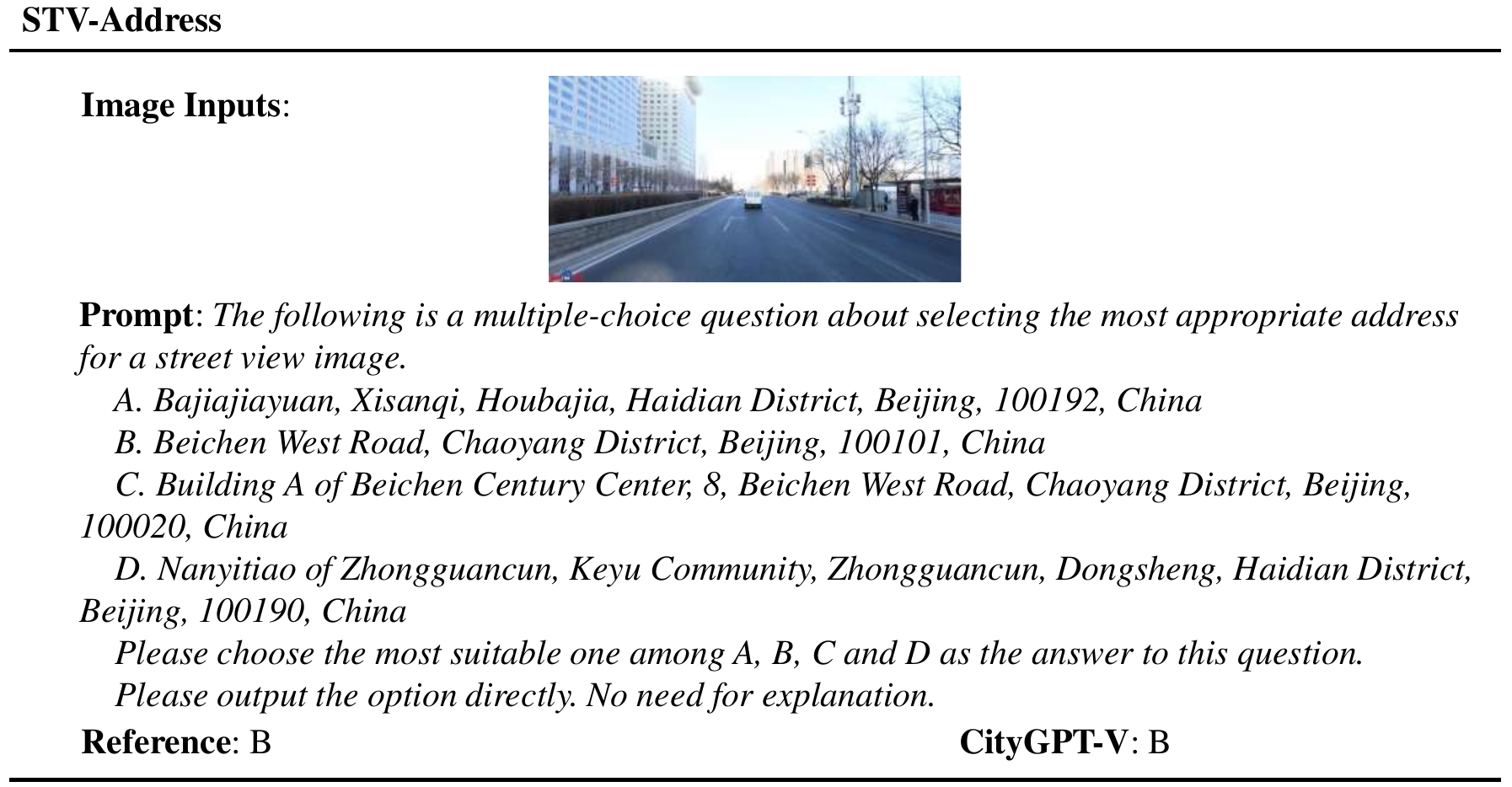}

   \caption{Example of a STV-Address task.}
   \label{fig:success_case_STV_address}
\end{figure}

\textbf{STV-Address.} This task provides a street view image and needs a model to speculate the most probable address this image was taken. Figure \ref{fig:success_case_STV_address} is an example.

\begin{figure}[t]
  \centering

   \includegraphics[width=1\linewidth]{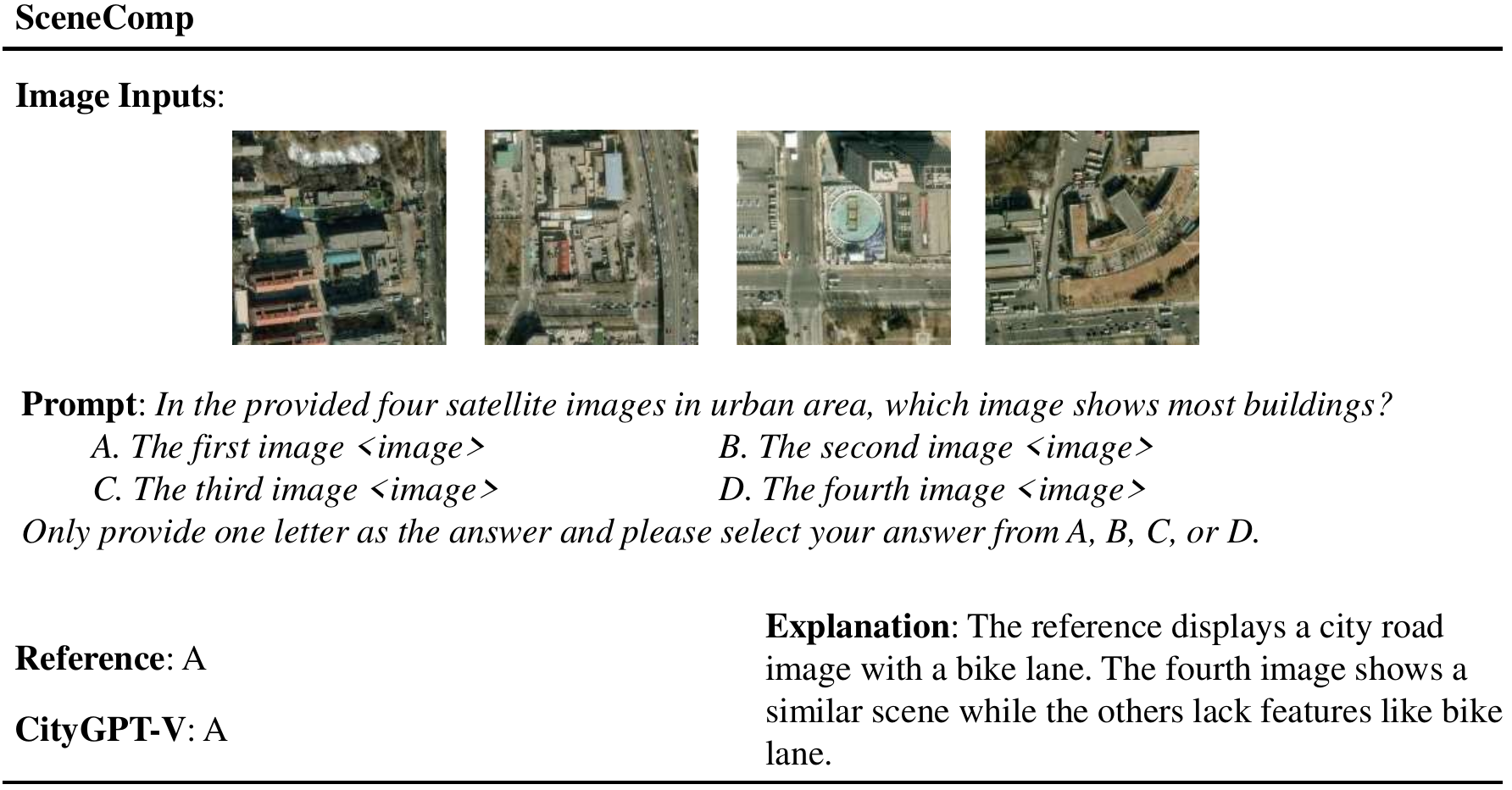}

   \caption{An example of a SceneComp task.}
   \label{fig:success_case_SceneComp}
\end{figure}

\textbf{SceneComp.} This task provides four satellite remote sensing images and prompts the model to choose the one with the most number of buildings. An instance is shown in Figure \ref{fig:success_case_SceneComp}.

\begin{figure}[t]
  \centering

   \includegraphics[width=1\linewidth]{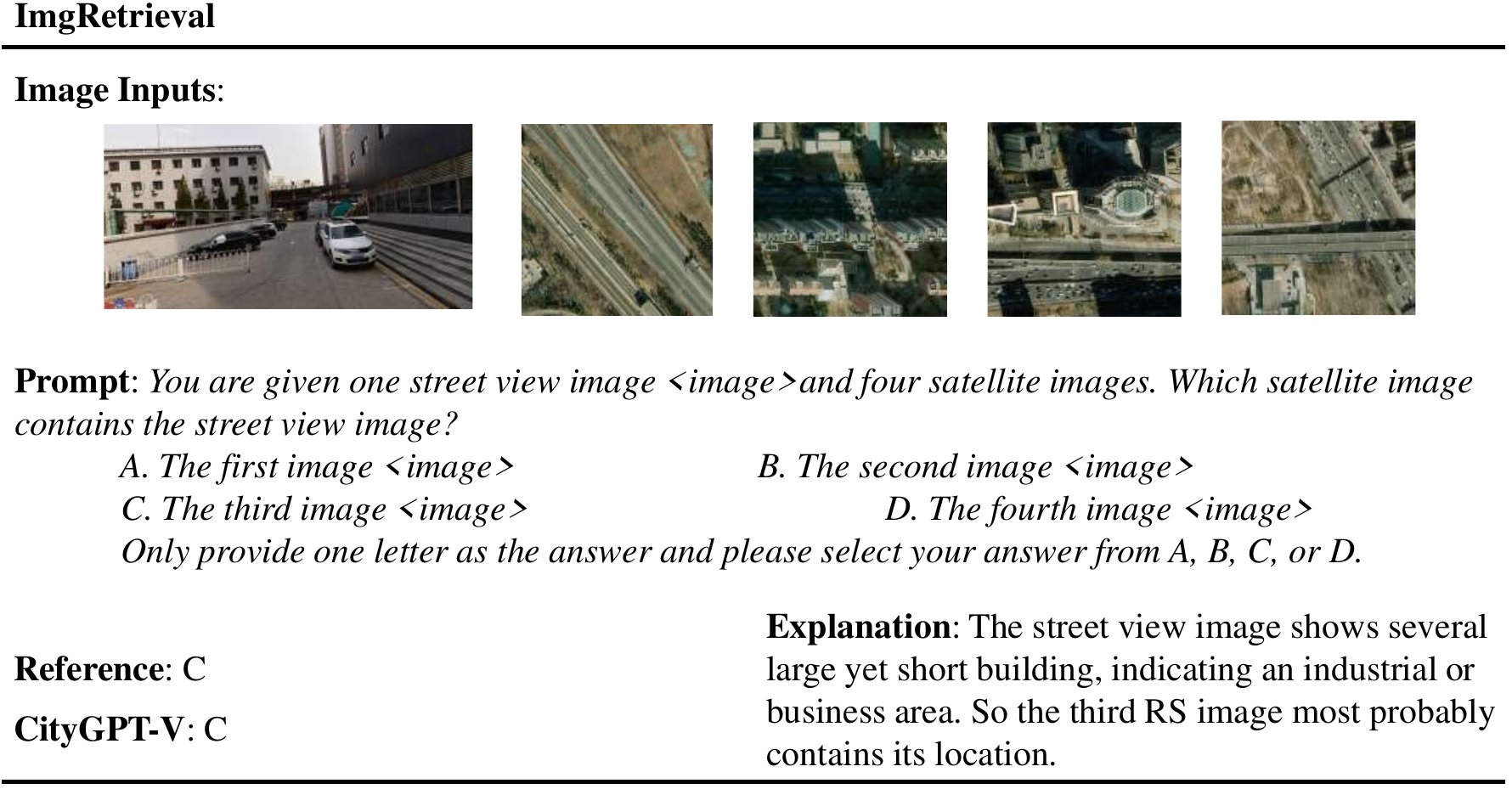}

   \caption{An example of an ImagRetrieval task.}
   \label{fig:success_case_ImgRetrieval}
\end{figure}

\textbf{ImgRetrieval.} It evaluates capability to map a given street view image to the corresponding satellite image. An example is displayed in Figure \ref{fig:success_case_ImgRetrieval}.

\begin{figure}[t]
  \centering

   \includegraphics[width=1\linewidth]{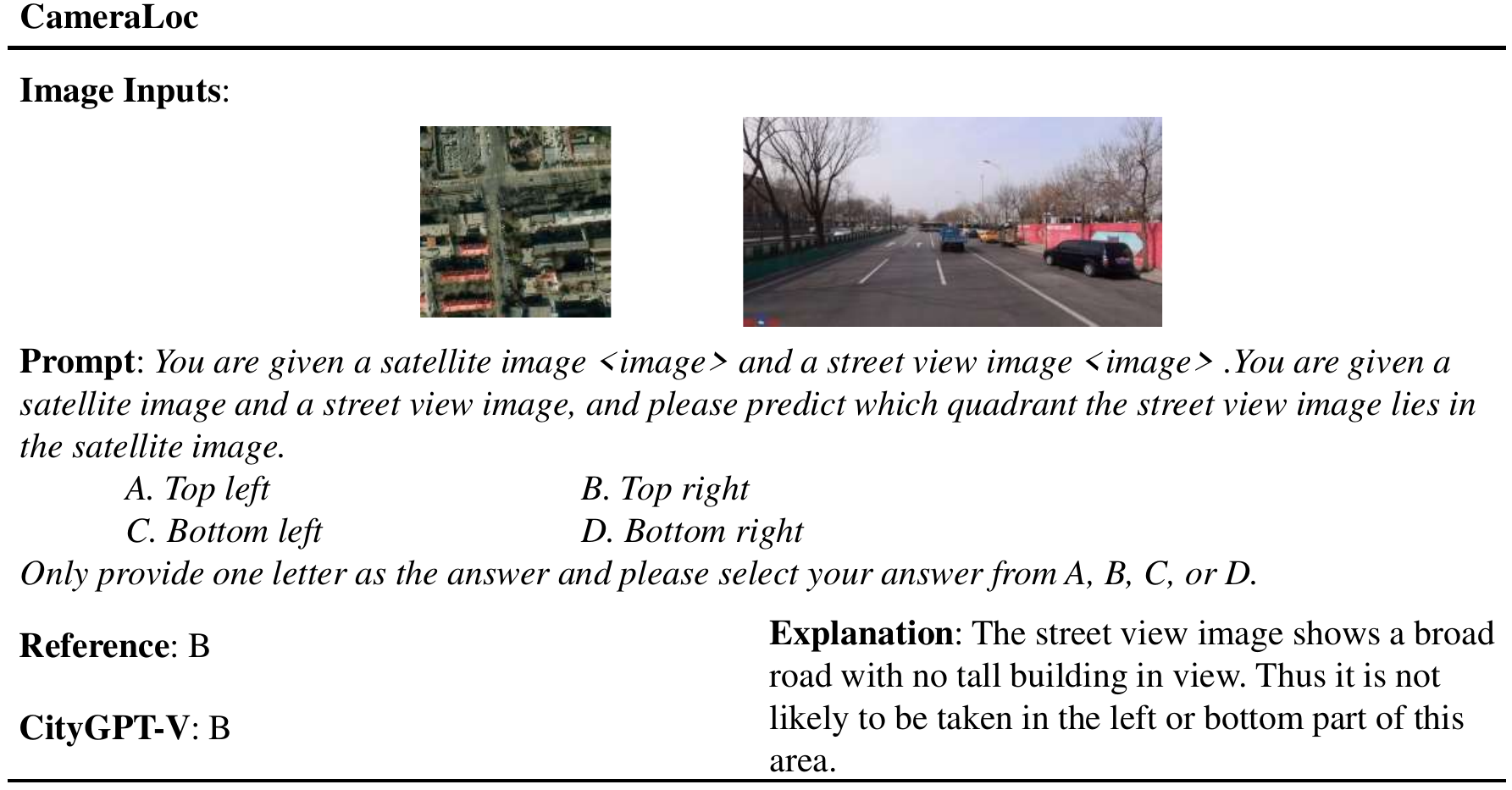}

   \caption{An example of a CameraLoc task.}
   \label{fig:success_case_CameraLoc}
\end{figure}

\textbf{CameraLoc.} It requires the model to infer which quadrant of a satellite image corresponds to the location where a given street view image was captured. An example is shown in Figure \ref{fig:success_case_CameraLoc}.

\section{Urban Instruction Data}
\begin{table}[h]
  \centering
  \setlength{\tabcolsep}{2pt} %
  \footnotesize %
\renewcommand\arraystretch{1.1} %
  \caption{Basic information of \CityGPTVData~on three cities.}
  \label{tab:training_data_statis}
  \resizebox{0.48\textwidth}{!}{
 \begin{tabular}{lcccc}
    \toprule
    \textbf{City} & \textbf{Category} & \textbf{Dataset} & \textbf{Instance} & \textbf{Rounds} \\
    \midrule
    \textbf{/} & General & ShareGPT,UltraChat,Open-Platypus & 19866 & 3.7\\
    \midrule
    \textbf{Beijing} & Location View Data & CityQA & 19271 & 1 \\
                     &                    & Location Address & 93246 & 1 \\
                     &                    & Landmark Details & 51130 & 1 \\
                     &                    & Image Description & 28798 & 1 \\
                     &                    & Cross Modality Reasoning & 2000 & 1 \\
                     \cmidrule{2-5}
                     & Trajectory View Data & Random Walk & 9001 & 1 \\
                     &                    & Real-World Trajectory & 98 & 1 \\
                     &                    & Visual Random Walk & 8936 & 1 \\
                     &                    & Vision-Language Navigation & 3000 & 1 \\
                     \cmidrule{2-5}
                     & Global View Data   & Image Content & 9315 & 1 \\
                     &                    & Location Address & 2777 & 1 \\
                     &                    & Landuse Inference & 3642 & 1 \\
                     &                    & Multiple SAT Comparison & 10114 & 1 \\
                     &                    & Cross-View Data & 77204 & 1 \\
                     &                    & Cross Modality Reasoning & 14977 & 1 \\
    \midrule
    \textbf{London}  & Location View Data & CityQA & 28934 & 1 \\
                     &                    & Location Address & 2172 & 1 \\
                     &                    & Landmark Details & 2372 & 1 \\
                     &                    & Image Description & 716 & 1 \\
                     &                    & Cross Modality Reasoning & 1286 & 1 \\
                     \cmidrule{2-5}
                     & Trajectory View Data & Random Walk & 16524 & 1 \\
                     &                    & Real-World Trajectory & 98 & 1 \\
                     &                    & Visual Random Walk & 13412 & 1 \\
                     &                    & Vision-Language Navigation & 3000 & 1 \\
                     \cmidrule{2-5}
                     & Global View Data   & Image Content & 3853 & 1 \\
                     &                    & Location Address & 882 & 1 \\
                     &                    & Landuse Inference & 4332 & 1 \\
                     &                    & Multiple SAT Comparison & 4500 & 1 \\
                     &                    & Cross-View Data & 2172 & 1 \\
                     &                    & Cross Modality Reasoning & 5758 & 1 \\
    \midrule
    \textbf{New York} & Location View Data & CityQA & 25413 & 1 \\
                      &                    & Location Address & 94886 & 1 \\
                      &                    & Landmark Details & 50404 & 1 \\
                      &                    & Image Description & 24529 & 1 \\
                      &                    & Cross Modality Reasoning & 2012 & 1 \\
                                          
                      \cmidrule{2-5}
                      & Trajectory View Data & Random Walk & 12277 & 1 \\
                      &                    & Real-World Trajectory & 98 & 1 \\
                      &                    & Visual Random Walk & 12229 & 1 \\
                      &                    & Vision-Language Navigation & 3000 & 1 \\
                      \cmidrule{2-5}
                      & Global View Data    & Image Content & 18368 & 1 \\
                      &                    & Location Address & 5113 & 1 \\
                      &                    & Landuse Inference & 17899 & 1 \\
                      &                    & Multiple SAT Comparison & 22020 & 1 \\
                      &                    & Cross-View Data & 94886 & 1 \\
                      &                    & Cross Modality Reasoning & 23603 & 1 \\
    \bottomrule
  \end{tabular}}
\end{table}

\begin{figure*}[]
\centering

\begin{tcolorbox}

\textbf{User:} \textless image \textgreater \textbackslash n Could you provide a description of the content shown in the arial image.

\tcblower

\centering
\begin{minipage}{0.45\textwidth}
    \centering
    \includegraphics[width=\linewidth]{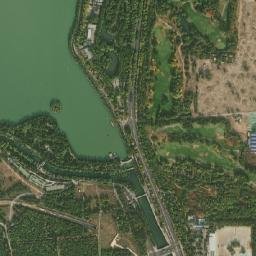} %
\end{minipage}

\vspace{10pt}
\rule[0mm]{\textwidth}{0.2mm} \\
\vspace{10pt}

\raggedright
\textbf{Assistant:} In the 256x256 pixel area, the road network is primarily composed of two main roads: the trunk road North Roud of Landianchang, which runs from [179, 249] to [178, 247] in a southward direction, and the service road Road of Kunminghu, extending from [127, 124] to [84, 16], which travels in a generally southwest direction. Additionally, there is a tertiary road Road of Kunminghu that connects [131, 123] to [127, 124], running parallel to the service road, and a secondary road North Roud of Landianchang that stretches from [176, 250] to [172, 248], also heading southward. The area features notable points of interest (POIs), including tickets located at [137, 175] and attractions at [88, 140]. The distribution of roads indicates a concentration of infrastructure towards the southern part of the region, particularly along North Roud of Landianchang, which is flanked by various POIs, while Road of Kunminghu serves as a key access route with a notable presence of attractions nearby, suggesting a vibrant area for visitors.

\end{tcolorbox}
\caption{An example of global view training instances of Image Content.}
\label{fig:data:basic}
\end{figure*}

\begin{figure*}[]
\centering
\label{train_example_basic_sat_descp}
\begin{tcolorbox}

\textbf{User:} \textless image \textgreater \textbackslash n The following is a multiple-choice question about selecting the most possible landuse type in the region of a satellite image.\\
A. Industrial\\
B. Residential\\
C. Meadow\\
D. Forest\\
Please choose the most suitable one among A, B, C and D as the answer to this question. \\
Please output the option directly. No need for explaination.
\tcblower

\centering
\begin{minipage}{0.45\textwidth}
    \centering
    \includegraphics[width=\linewidth]{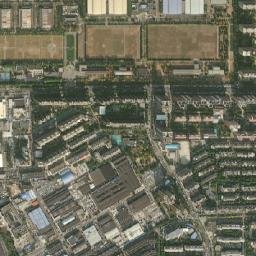} %
\end{minipage}

\vspace{10pt}
\rule[0mm]{\textwidth}{0.2mm} \\
\vspace{10pt}

\raggedright
\textbf{Assistant:} B

\end{tcolorbox}
\caption{An example of global view training instances of Landuse Inference.}
\end{figure*}

\begin{figure*}[]
\centering
\begin{tcolorbox}

\textbf{User:} You are given one street view image \textless image\textgreater and four satellite images. Which satellite image shows the surroundings of the street view image?\\
A. The first image \textless image\textgreater \\          
B. The second image \textless image\textgreater \\
C. The third image \textless image\textgreater \\
D. The fourth image \textless image\textgreater \\
Only provide one letter as the answer and please select your answer from A, B, C, or D.

\tcblower

\centering
\begin{minipage}{0.5\textwidth}
    \centering
    \includegraphics[width=\linewidth]{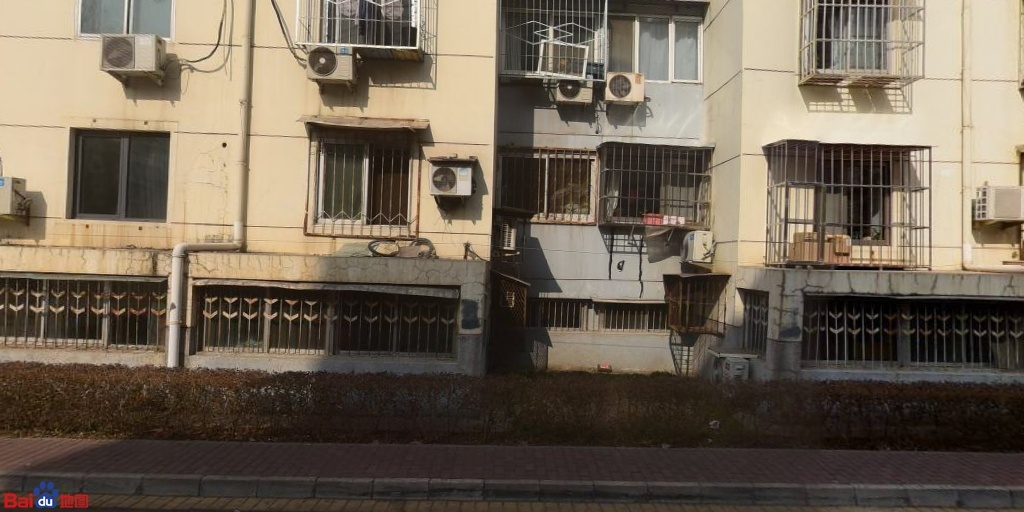} %
\end{minipage}
\hfill
\begin{minipage}{0.3\textwidth}
    \centering
    \includegraphics[width=\linewidth]{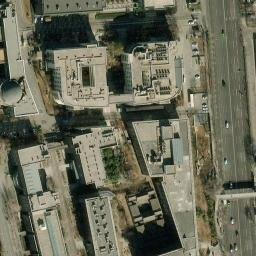} %
\end{minipage}
\hfill

\vspace{10pt} %

\begin{minipage}{0.3\textwidth}
    \centering
    \includegraphics[width=\linewidth]{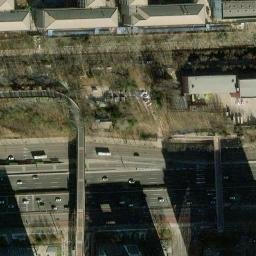} %
\end{minipage}
\hfill
\begin{minipage}{0.3\textwidth}
    \centering
    \includegraphics[width=\linewidth]{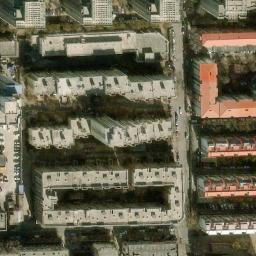} %
\end{minipage}
\hfill
\begin{minipage}{0.3\textwidth}
    \centering
    \includegraphics[width=\linewidth]{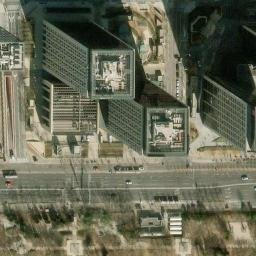} %
\end{minipage}

\vspace{10pt}
\rule[0mm]{\textwidth}{0.2mm} \\
\vspace{10pt}

\raggedright

\textbf{Assistant:} C

\end{tcolorbox}
\caption{An example of global view training instances of Multiple SAT Comparison.}
\label{fig:data:multi}
\end{figure*}

Table~\ref{tab:training_data_statis} provides detailed statistics of \CityGPTVData~across three cities, while Table~\ref{tab:orign-data} presents the detailed statistics of the raw data used to construct \CityGPTVData. Additionally, we present representative examples of our urban instruction data, as shown in Figure~\ref{fig:data:basic} to Figure~\ref{fig:data:multi}.

\begin{figure*}[]
\centering
\begin{tcolorbox}

\textbf{User:} \textless image \textgreater \textbackslash n You are provided a 256*256 satellite image. What is the landuse type in region [[32, 136], [33, 167], [39, 167], [101, 133], [99, 127], [90, 131], [86, 121], [88, 114], [74, 111], [32, 136]]?
\tcblower

\centering
\begin{minipage}{0.45\textwidth}
    \centering
    \includegraphics[width=\linewidth]{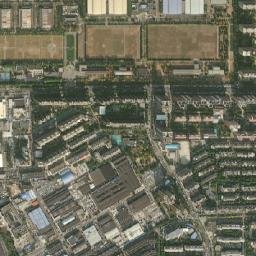} %
\end{minipage}

\vspace{10pt}
\rule[0mm]{\textwidth}{0.2mm} \\
\vspace{10pt}

\raggedright
\textbf{Assistant:} Residential

\end{tcolorbox}
\caption{An example of global view training instances of Landuse Inference.}
\end{figure*}

\begin{figure*}[]
\centering
\begin{tcolorbox}

\textbf{User:} \textless image \textgreater \textbackslash n What is the detailed address of the provided street view image?
\tcblower

\centering
\begin{minipage}{0.45\textwidth}
    \centering
    \includegraphics[width=\linewidth]{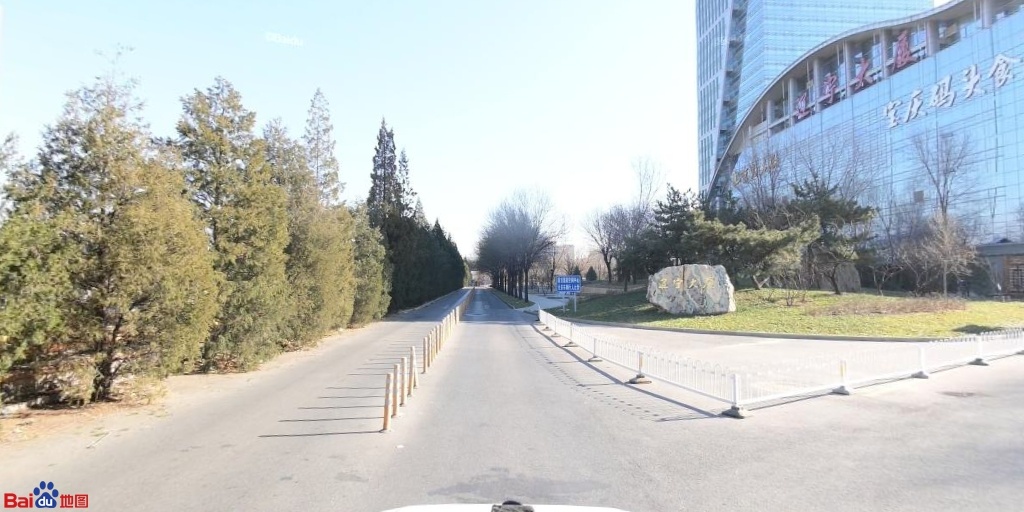} %
\end{minipage}

\vspace{10pt}
\rule[0mm]{\textwidth}{0.2mm} \\
\vspace{10pt}

\raggedright
\textbf{Assistant:} Tsinghuayuan, Xueyuan Road, Dongsheng, Haidian District, Beijing, 100190, China

\end{tcolorbox}
\caption{An example of local view training instances of Location Address.}
\end{figure*}

\begin{figure*}[]
\centering
\begin{tcolorbox}

\textbf{User:} \textless image \textgreater \textbackslash n What is shown in this street view image? And what related pois could be estimated nearby
\tcblower

\centering
\begin{minipage}{0.45\textwidth}
    \centering
    \includegraphics[width=\linewidth]{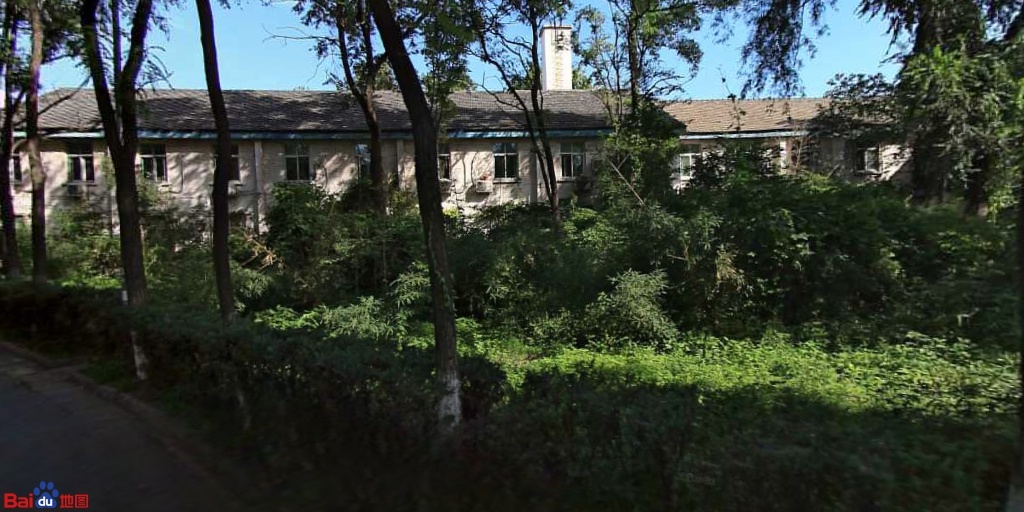} %
\end{minipage}

\vspace{10pt}
\rule[0mm]{\textwidth}{0.2mm} \\
\vspace{10pt}

\raggedright
\textbf{Assistant:} The image depicts a building partially obscured by lush greenery. The structure appears to be a single-story building with a flat roof, characterized by a light-colored exterior that suggests a concrete or stucco material. The roof is lined with a blue trim, which adds a subtle contrast to the overall appearance.\\
Surrounding the building, dense foliage and various plants create a vibrant green backdrop, indicating a well-established natural environment. The vegetation is thick, with a mix of taller trees and lower shrubs, suggesting a sense of seclusion. The building is positioned towards the center of the image, with trees flanking its sides, while a pathway or road runs along the left side, leading towards the viewer. \\
The spatial relationship between the building and the surrounding greenery is notable; the building is somewhat hidden by the dense plants, which extend close to its walls, creating a harmonious blend of man-made and natural elements. The sunlight filters through the leaves, casting dappled shadows on the ground, enhancing the tranquil atmosphere of the scene. Overall, the image conveys a sense of quietude and integration with nature.

\end{tcolorbox}
\caption{An example of local view training instances of Image Description.}
\end{figure*}

\begin{figure*}[]
\centering
\begin{tcolorbox}

\textbf{User:} \textless image \textgreater \textbackslash n What is the landmark in this image? If there is no landmark, please say 'None'.
\tcblower

\centering
\begin{minipage}{0.45\textwidth}
    \centering
    \includegraphics[width=\linewidth]{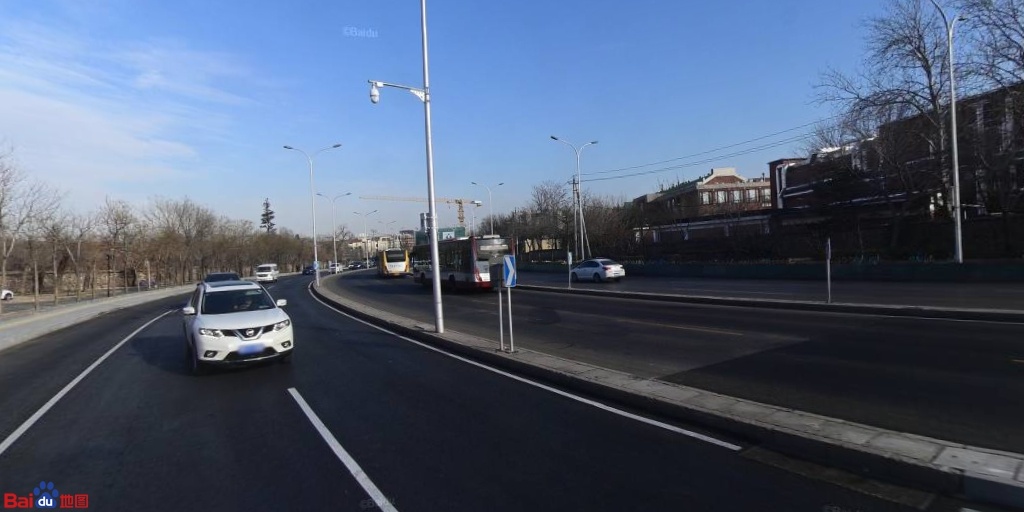} %
\end{minipage}

\vspace{10pt}
\rule[0mm]{\textwidth}{0.2mm} \\
\vspace{10pt}

\raggedright
\textbf{Assistant:} In the street view image, a notable landmark is visible in the background: a pagoda, which is likely part of the nearby Changchun Garden. The pagoda stands tall among the trees, providing a distinctive silhouette against the sky, and serves as a navigational reference point for the surrounding area. The road curves gently around the landmark, with vehicles traveling in both directions, while the nearby buildings, including educational institutions like the Medical College and Pharmacy College, create a backdrop that emphasizes the pagoda's prominence in the landscape. This combination of natural and architectural features highlights the cultural significance of the pagoda within the urban environment

\end{tcolorbox}
\caption{An example of local view training instances of Landmark Details.}
\end{figure*}

\begin{figure*}[]
\centering
\begin{tcolorbox}

\textbf{User:} You are given a satellite image \textless image\textgreater and a street view image \textless image\textgreater, can you predict which quadrant the street view image lies in the satellite image?\\
A. Top left\\
B. Top right\\
C. Bottom left\\
D. Bottom right\\
Please provide a detailed reasoning for your answer. And give me one letter as your answer in the last sentence.

\tcblower

\centering
\begin{minipage}{0.45\textwidth}
    \centering
    \includegraphics[width=\linewidth]{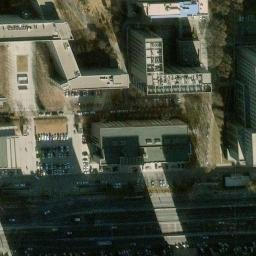} %
\end{minipage}
\hfill
\begin{minipage}{0.45\textwidth}
    \centering
    \includegraphics[width=\linewidth]{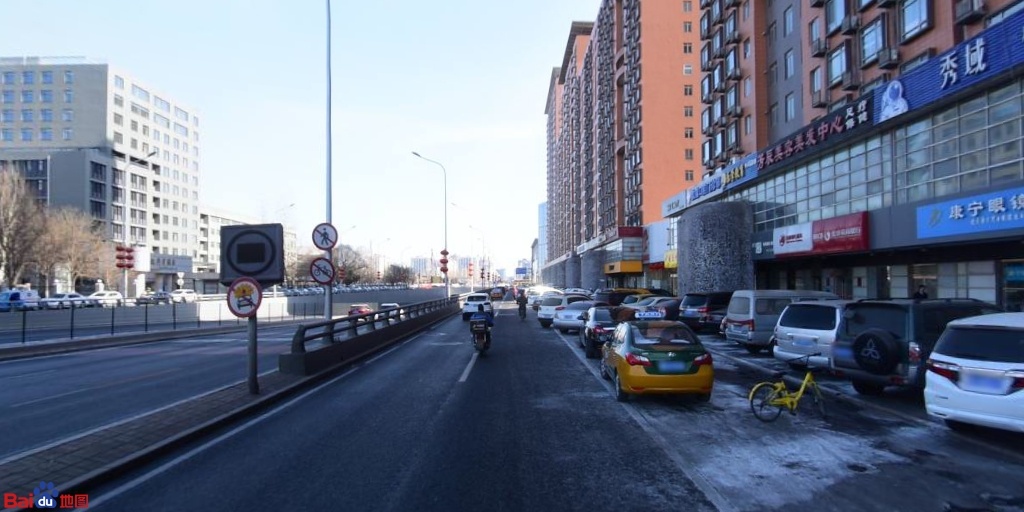} %
\end{minipage}

\vspace{10pt}
\rule[0mm]{\textwidth}{0.2mm} \\
\vspace{10pt}

\raggedright
\textbf{Assistant:} Step 1: Identify the satellite view image's address: The satellite image is taken in the Chaoyang District of Beijing, featuring a well-organized 5x5 grid of notable locations. To the north is the Olympic Village Science Park, a central site surrounded by various establishments. Directly east along Datun Road is the National Zoological Museum of China, enhancing the area's scientific theme. Moving south on Datun Road, you'll find China Unicom, a key telecommunications provider, and the vibrant The Red Parrot Image Agency, adding cultural flair. Datun Road connects these points, facilitating easy navigation. This area blends educational, cultural, and commercial establishments, all within close proximity.\\
Step 2: Identify the street view image's address: The street view image is taken at The Red Parrot Image Agency, \# 50, Datun Road, Chaoyang District, Beijing, 100101, China.\\
Step 3: Determine the quadrant where the street view image lies in the satellite view image: The street view image is located in quadrant C of the satellite view image."

\end{tcolorbox}
\caption{An example of local view training instances of Cross Modality Reasoning.}
\end{figure*}

\begin{figure*}[]
\centering
\begin{tcolorbox}
\textbf{Assistant:}  You are given one street view image \textless image\textgreater and four satellite images. Which satellite image shows the surroundings of the street view image?\\
A. The first image \textless image\textgreater \\
B. The second image \textless image\textgreater \\
C. The third image \textless image\textgreater \\            
D. The fourth image \textless image\textgreater \\
Please provide a detailed reasoning for your answer. And give me one letter as your answer in the last sentence."

\tcblower

\centering
\begin{minipage}{0.5\textwidth}
    \centering
    \includegraphics[width=\linewidth]{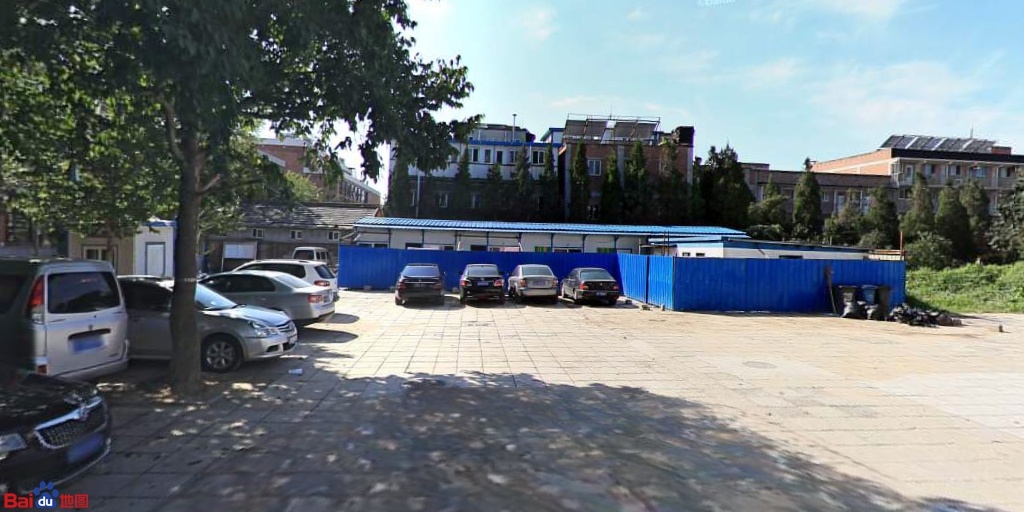} %
\end{minipage}
\hfill
\begin{minipage}{0.3\textwidth}
    \centering
    \includegraphics[width=\linewidth]{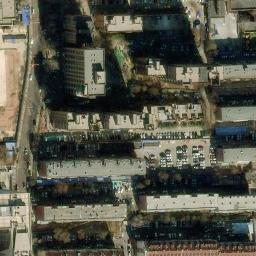} %
\end{minipage}
\hfill

\vspace{10pt} %

\begin{minipage}{0.3\textwidth}
    \centering
    \includegraphics[width=\linewidth]{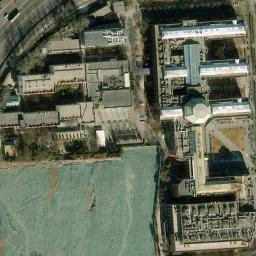} %
\end{minipage}
\hfill
\begin{minipage}{0.3\textwidth}
    \centering
    \includegraphics[width=\linewidth]{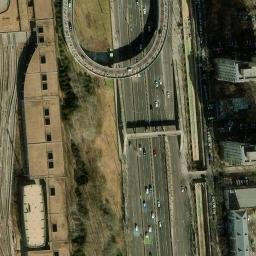} %
\end{minipage}
\hfill
\begin{minipage}{0.3\textwidth}
    \centering
    \includegraphics[width=\linewidth]{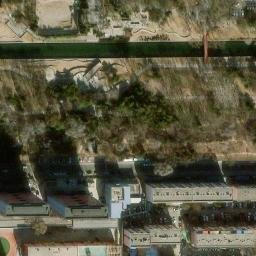} %
\end{minipage}

\vspace{10pt}
\rule[0mm]{\textwidth}{0.2mm} \\
\vspace{10pt}

\raggedright

\textbf{Assistant:} Step 1: Describe the street view image: The image shows a parking area with various vehicles parked in a semi-organized manner. In the foreground, cars of different colors, including silver, black, and white, are parked closely on a paved surface made of large rectangular tiles, indicating moderate occupancy. To the right, a prominent blue fence encloses a structure, likely a building or series of buildings, contrasting with nearby lush trees. The background features multi-storied buildings made of brick and concrete, with visible windows, suggesting residential or commercial use. The parking area is in front of these buildings, with cars facing the blue enclosure. The short distance between the cars and the fence creates a compact environment. The clear sky suggests a bright day, enhancing visibility.\\
Step 2: Recall the address of the street view image: The image is taken at the School of Medicine, Jinchun Road, Qinghuayuan, Xiwacun, Haidian District, Beijing, 100084, China.\\
Step 3: Recall the addresses of each possible satellite view image: \\
... \\
Step 4: Reason the corresponding satellite view image: The satellite view image corresponding to the street view image is B

\end{tcolorbox}
\caption{An example of global view training instances of Cross Modality Reasoning.}
\end{figure*}

\begin{figure*}[]
\centering
\begin{tcolorbox}

\textbf{User:} \textless image \textgreater \textbackslash n What is the address of this location? Please think step by step and provide a detailed reasoning.
\tcblower

\centering
\begin{minipage}{0.45\textwidth}
    \centering
    \includegraphics[width=\linewidth]{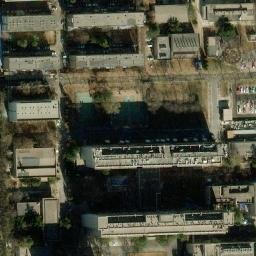} %
\end{minipage}

\vspace{10pt}
\rule[0mm]{\textwidth}{0.2mm} \\
\vspace{10pt}

\raggedright
\textbf{Assistant:} Step 1: Identify the city name. According to the satellite view image, this image was taken in Beijing.\\
Step 2: Extract the location's features. From the image, I can observe the following features: In the 256x256 pixel area, the roads are primarily oriented in a north-south and east-west direction, with significant intersections facilitating access to various points of interest (POIs). Notably, the coordinates [116, 40] host both research institutes and universities, indicating a concentration of educational and research facilities in this region. These POIs are strategically located alongside the main north-south road that runs through the area, enhancing their accessibility for students and researchers alike. The distribution of roads in this region is fairly balanced, with a few major thoroughfares connecting key locations. The concentration of specific POIs, particularly educational institutions, suggests that this area serves as a hub for academic and research activities. Overall, the layout of roads and the placement of POIs create a vibrant environment conducive to learning and innovation.\\
Step 3: Identify the location's address. Based on my observation and knowledge of this region, the area is situated in Haidian District, Beijing, characterized by a 5x5 layout that features several prominent institutions. To the north, the Institute of Semiconductors, Chinese Academy of Sciences, occupies multiple locations, creating a significant presence in the region. Directly to the south, Beijing Forestry University is also located, providing an educational backdrop to the area. The main thoroughfare, Beilin Ginkgo Avenue, runs through the region, connecting these institutions and facilitating easy access. The Xueyuanlu Subdistrict, where this area is found, enhances the accessibility of the institutions, with Bajia Village nearby, further enriching the local community. The proximity of these educational and research institutions fosters a collaborative environment, making this area a hub for academic and scientific endeavors.

\end{tcolorbox}
\caption{An example of global view training instances of Image Content.}
\end{figure*}

\begin{figure*}[]
\centering
\begin{tcolorbox}

\textbf{User:} \textless image \textgreater \textbackslash n Please count the number of buildings in this satellite image and provide a detailed reasoning for your answer.
\tcblower

\centering
\begin{minipage}{0.45\textwidth}
    \centering
    \includegraphics[width=\linewidth]{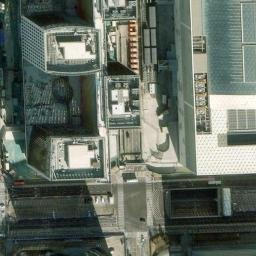} %
\end{minipage}

\vspace{10pt}
\rule[0mm]{\textwidth}{0.2mm} \\
\vspace{10pt}

\raggedright
\textbf{Assistant:} Step 1: Identify the city name: According to the satellite view image, this image was likely taken in Beijing. \\
Step 2: Describe the location's features: From the image, I can observe the following features: In the 256x256 pixel area, the roads are arranged to facilitate access to various points of interest (POIs) and ensure smooth traffic flow. The tertiary road, Tianchen West Road, runs vertically from [137, 188] to [137, 171], providing a north-south route that connects to nearby POIs, including cars located at [47, 155]. The secondary road, Datun Road, extends horizontally from [174, 211] to [221, 209], serving as a major thoroughfare likely supporting higher traffic volumes and is near cafes located at [55, 148] and [61, 142]. Additionally, another segment of the tertiary road, Datun Road, stretches from [162, 187] to [0, 193], indicating a more extensive network that enhances connectivity across the region. The concentration of cafes near the secondary road suggests a vibrant social scene, while the presence of cars indicates active transportation throughout the area, particularly along these key roads. Overall, the layout of the roads and the distribution of POIs create a dynamic environment that caters to both vehicular movement and local amenities.\\
Step 3: Provide the location's address: Based on my observation and knowledge of this region, the area described is located in the Chaoyang District of Beijing, characterized by a well-organized 5x5 layout. To the north, you will find Beichen West Road, a prominent thoroughfare, while to the south, Datun North Road runs parallel, providing easy access to the surrounding locations. The eastern section is defined by Tianchen West Road, which intersects with various points of interest, including the Beichen Century Center, situated at the northwest corner. This center is flanked by additional buildings along Beichen West Road, creating a bustling environment. Adjacent to these landmarks, Datun Road runs through the area, connecting several establishments and enhancing the region's accessibility. The overall layout ensures that each location is conveniently reachable, fostering a vibrant community atmosphere.\\
Step 4: Count the number of buildings or POIs: There are 5 buildings in this image."

\end{tcolorbox}
\caption{An example of global view training instances of Cross Modality Reasoning.}
\end{figure*}

\begin{figure*}[]
\centering
\begin{tcolorbox}

\textbf{User:} \textless image \textgreater \textbackslash n What is the address of this location? Please think step by step and provide a detailed reasoning.
\tcblower

\centering
\begin{minipage}{0.45\textwidth}
    \centering
    \includegraphics[width=\linewidth]{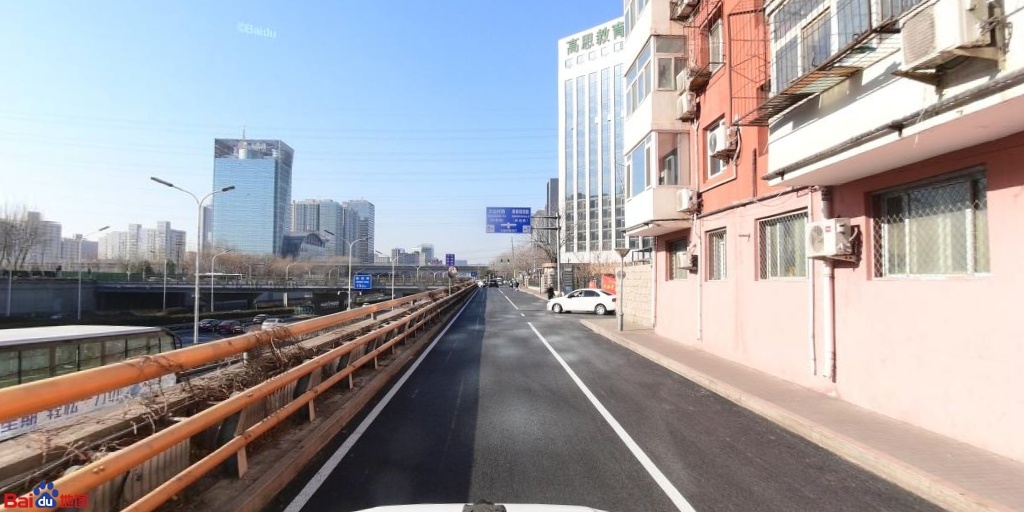} %
\end{minipage}

\vspace{10pt}
\rule[0mm]{\textwidth}{0.2mm} \\
\vspace{10pt}

\raggedright
\textbf{Assistant:} Step 1: Describe the street view image: This is a street view image depicting a city street scene with a wide, well-maintained road extending into the distance. On the left side, a prominent orange guardrail runs parallel to the road, suggesting a separation from a lower area, possibly a sidewalk or a bus stop. The road is dark asphalt, reflecting the bright sunlight, indicating a clear day. To the right, a series of buildings are visible, showcasing a mix of architectural styles. The building closest to the viewer is painted in a muted red hue and features several air conditioning units mounted on its exterior walls. This building appears to be several stories tall, with large windows that allow for natural light. Further along, a taller, modern glass building stands out, its reflective surface capturing the sunlight and contrasting with the surrounding structures. This building is flanked by additional high-rise buildings in the background, creating a skyline that suggests a bustling urban environment. In the distance, road signs are visible, indicating directions or information for drivers, while the road continues to stretch ahead, leading towards the horizon. The overall scene conveys a sense of an active city, with a combination of modern architecture and well-maintained infrastructure. The clear blue sky above enhances the vibrant atmosphere of the setting.\\
Step 2: Tell the city name: According to the street view image, this is probably in Beijing.\\
Step 3: Extract the location's features around: The street view image is taken in a region with the following features: Beijing Haidian District Vocational School, Haidian District Vocational Technical School East Campus, Hongyun Laobeiping, Khan Baba, Mi Da Wan, and the Affiliated High School of China University of Geosciences.\\
Step 4: Tell the location's address: Based on my observation and knowledge about this region, the address is Xueyuanlu Subdistrict, Dongsheng, Haidian District, Beijing, 100083, China."
            
\end{tcolorbox}
\caption{An example of local view training instances of Cross Modality Reasoning.}
\end{figure*}

\begin{figure*}[]
\centering
\begin{tcolorbox}

\textbf{User:} \textless image \textgreater \textbackslash n The following is a multiple-choice question about selecting the most appropriate address for a satellite image.\\
A. The area is characterized by a well-organized layout, featuring a 5x5 grid that offers a variety of attractions and roadways. To the north, the prominent Yuan Dynasty Capital City Wall Ruins Park serves as a significant landmark, providing a historical backdrop and recreational space. Adjacent to this park, Beitou City West Road runs east to west, connecting various points of interest within the vicinity. To the east ...\\
B. The area is situated in the Huayuanlu Subdistrict of Haidian District, Beijing, characterized by a well-defined square layout divided into a 5x5 arrangement. To the north, the region is bordered by a series of residential complexes, including the prominent Beiji Temple Residential Area, which occupies several positions within the grid. To the south ...\\
C. The area is characterized by a well-organized 5x5 layout, with the northern section primarily featuring Wanquanhe Road, which runs through the heart of the neighborhood, providing easy access to various points of interest. To the east, ... \\
D. The area is located in the Huayuanlu Subdistrict of Haidian District, Beijing, characterized by a well-defined square layout that is predominantly residential. To the north, you will find a cluster of residential buildings known as Beiji Temple Residential Area, which extends down the western side of the area. Moving eastward, ...\\
Please choose the most suitable one among A, B, C and D as the answer to this question. \\
Please output the option directly. No need for explaination
\tcblower

\centering
\begin{minipage}{0.45\textwidth}
    \centering
    \includegraphics[width=\linewidth]{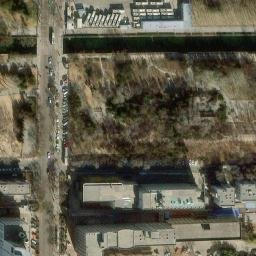} %
\end{minipage}

\vspace{10pt}
\rule[0mm]{\textwidth}{0.2mm} \\
\vspace{10pt}

\raggedright
\textbf{Assistant:} A

\end{tcolorbox}
\caption{An example of global view training instances of Location Address.}
\end{figure*}

\begin{figure*}[]
\centering
\begin{tcolorbox}

\textbf{User:} \textless image \textgreater \textbackslash n The following is a multiple-choice question about selecting the most appropriate address for a street view image.\\
A. G6 Auxiliary Road, Chaoyang District, Beijing, 100000, China\\
B. Old Summer Palace Park, 28, Qinghua West Road, Yanyuan, Haidian Town, Haidian District, Beijing, 100084, China\\
C. Piaoliang Sunshine Plaza, Datun, Chaoyang District, Beijing, 100012, China\\
D. G6 Auxiliary Road, Xueyuanlu Subdistrict, Tayuan Village, Haidian District, Beijing, 100191, China\\    Please choose the most suitable one among A, B, C and D as the answer to this question. \\
Please output the option directly. No need for explaination.
\tcblower

\centering
\begin{minipage}{0.45\textwidth}
    \centering
    \includegraphics[width=\linewidth]{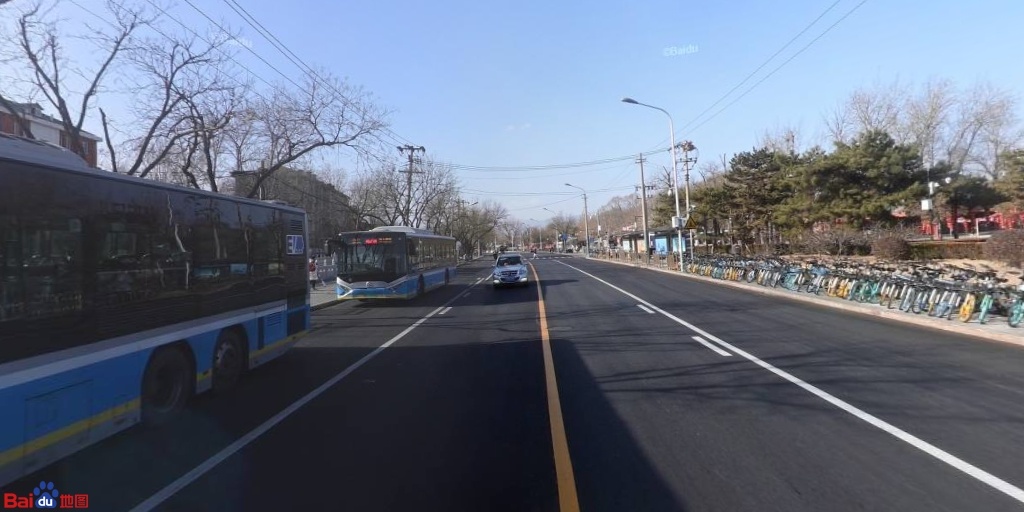} %
\end{minipage}

\vspace{10pt}
\rule[0mm]{\textwidth}{0.2mm} \\
\vspace{10pt}

\raggedright
\textbf{Assistant:} B

\end{tcolorbox}
\caption{An example of local view training instances of Location Address.}
\end{figure*}

\begin{figure*}[]
\centering
\begin{tcolorbox}

\textbf{User:} \textless image \textgreater \textbackslash n The following is a multiple-choice question about selecting the most possible nearby POIs(Place of Interests) or landmarks description in the region of a street view image.\\
A. Unpaved rural road.\\
B. Metal fence or wall.\\
C. Not identifiable.\\
D. Residential apartment complex.\\
Please choose the most suitable one among A, B, C and D as the answer to this question. \\
Please output the option directly. No need for explaination.
\tcblower

\centering
\begin{minipage}{0.45\textwidth}
    \centering
    \includegraphics[width=\linewidth]{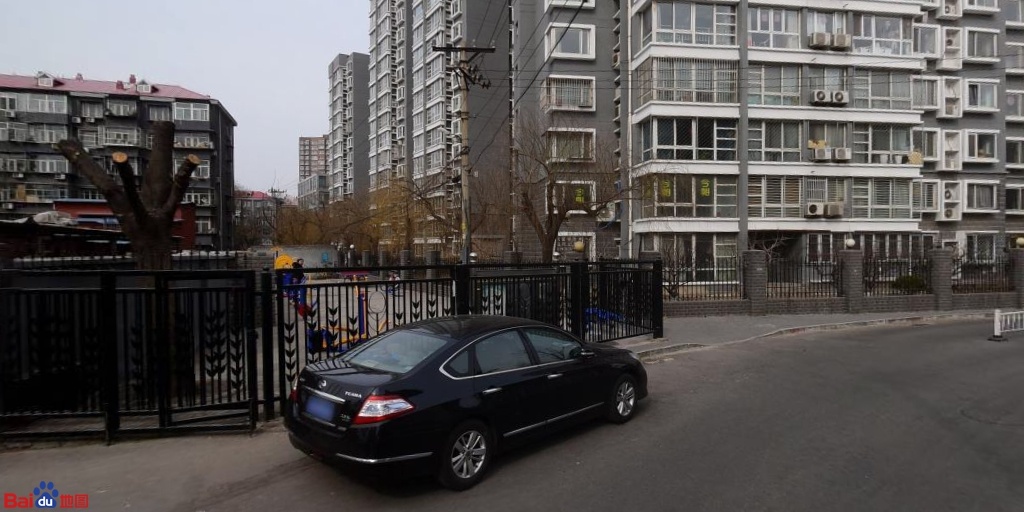} %
\end{minipage}

\vspace{10pt}
\rule[0mm]{\textwidth}{0.2mm} \\
\vspace{10pt}

\raggedright
\textbf{Assistant:} D

\end{tcolorbox}
\caption{An example of local view training instances of Landmark Details.}
\end{figure*}

\begin{figure*}[]
\centering
\begin{tcolorbox}

\textbf{User:} You are given a satellite image \textless image\textgreater and a street view image \textless image\textgreater, and please predict which quadrant the street view image lies in the satellite image.\\
A. Top left\\
B. Top right\\
C. Bottom left\\
D. Bottom right\\
Only provide one letter as the answer and please select your answer from A, B, C, or D."

\tcblower

\centering
\begin{minipage}{0.45\textwidth}
    \centering
    \includegraphics[width=\linewidth]{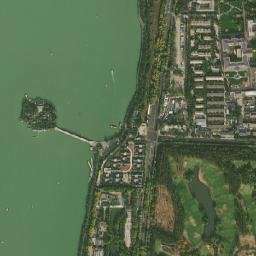} %
\end{minipage}
\hfill
\begin{minipage}{0.45\textwidth}
    \centering
    \includegraphics[width=\linewidth]{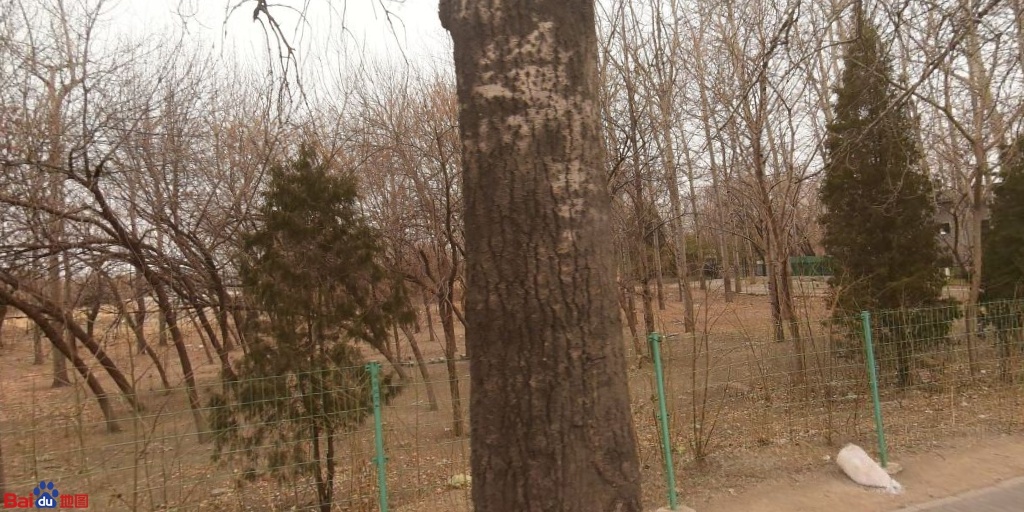} %
\end{minipage}

\vspace{10pt}
\rule[0mm]{\textwidth}{0.2mm} \\
\vspace{10pt}

\raggedright

\textbf{ASSITANT:} D

\end{tcolorbox}
\caption{An example of global view training instances of Cross View Data.}
\end{figure*}

\begin{figure*}[t]
  \centering
   \begin{subfigure}[b]{0.32\textwidth}   %
        \includegraphics[width=\textwidth, height=3cm]{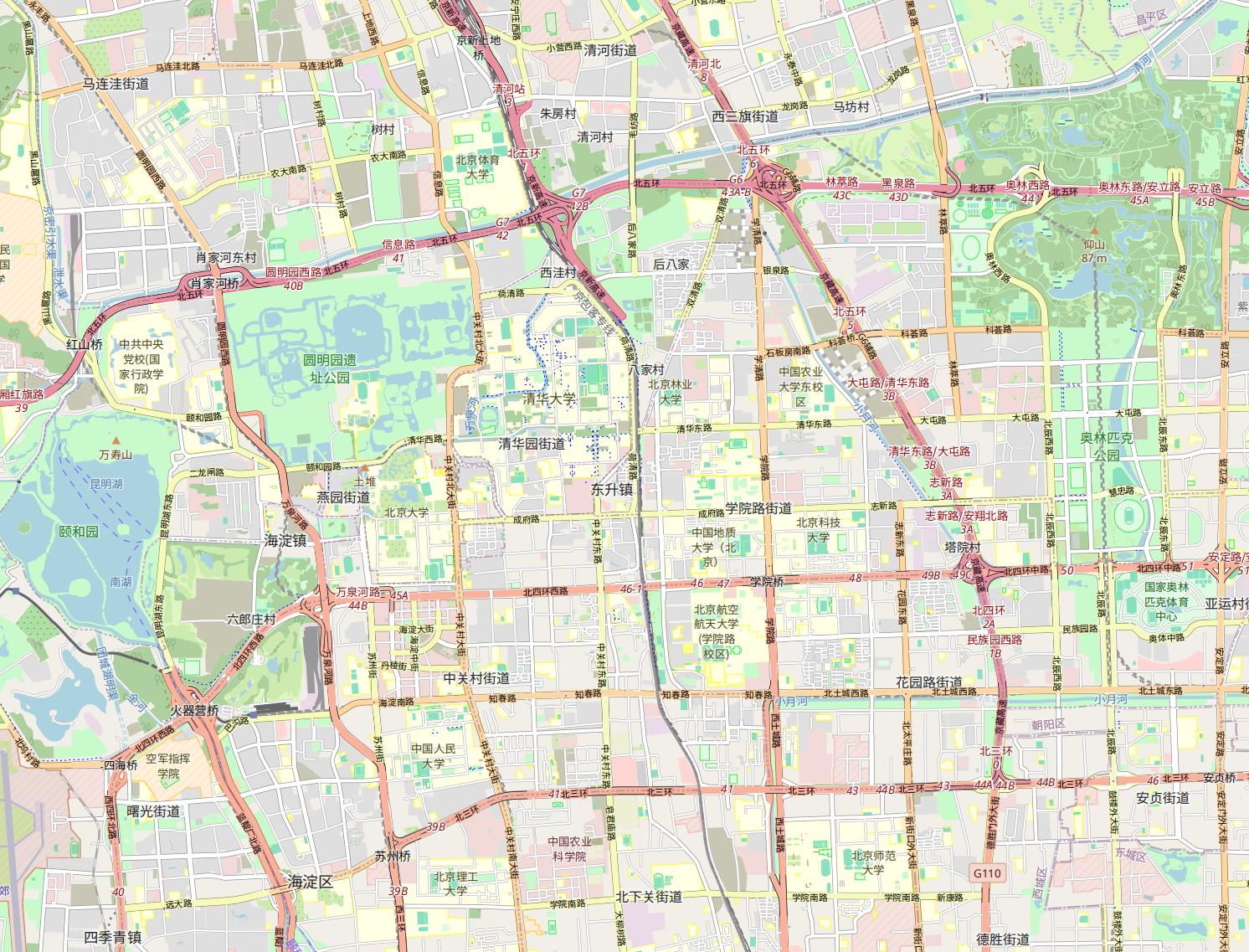} 
        \caption{Beijing}
        \label{fig:map_beijing}
    \end{subfigure}
    \hspace{1pt}  
    \begin{subfigure}[b]{0.32\textwidth}
        \includegraphics[width=\textwidth, height=3cm]{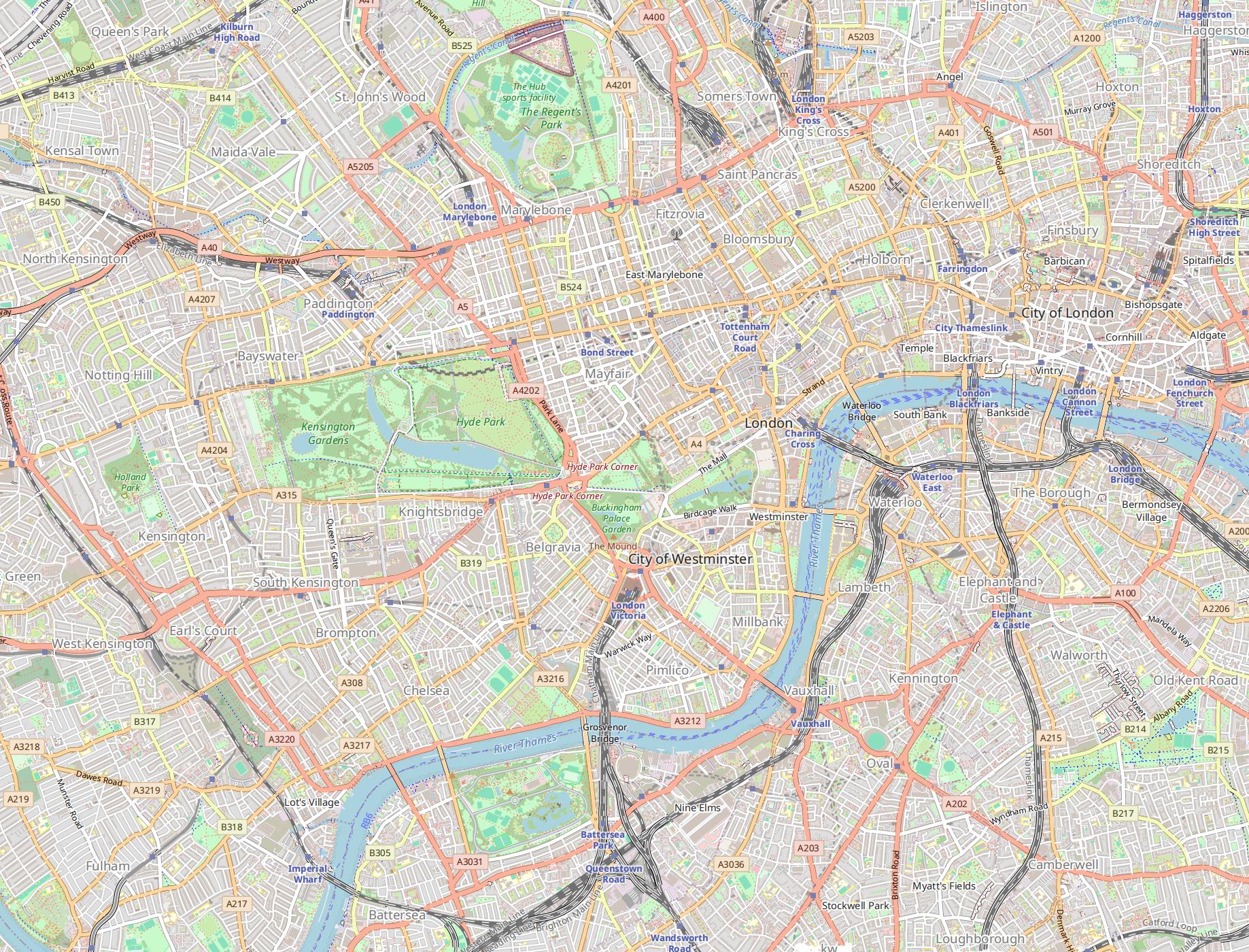} 
        \caption{London}
        \label{fig:map_london}
    \end{subfigure}
    \hspace{1pt}  
    \begin{subfigure}[b]{0.32\textwidth}
        \includegraphics[width=\textwidth, height=3cm]{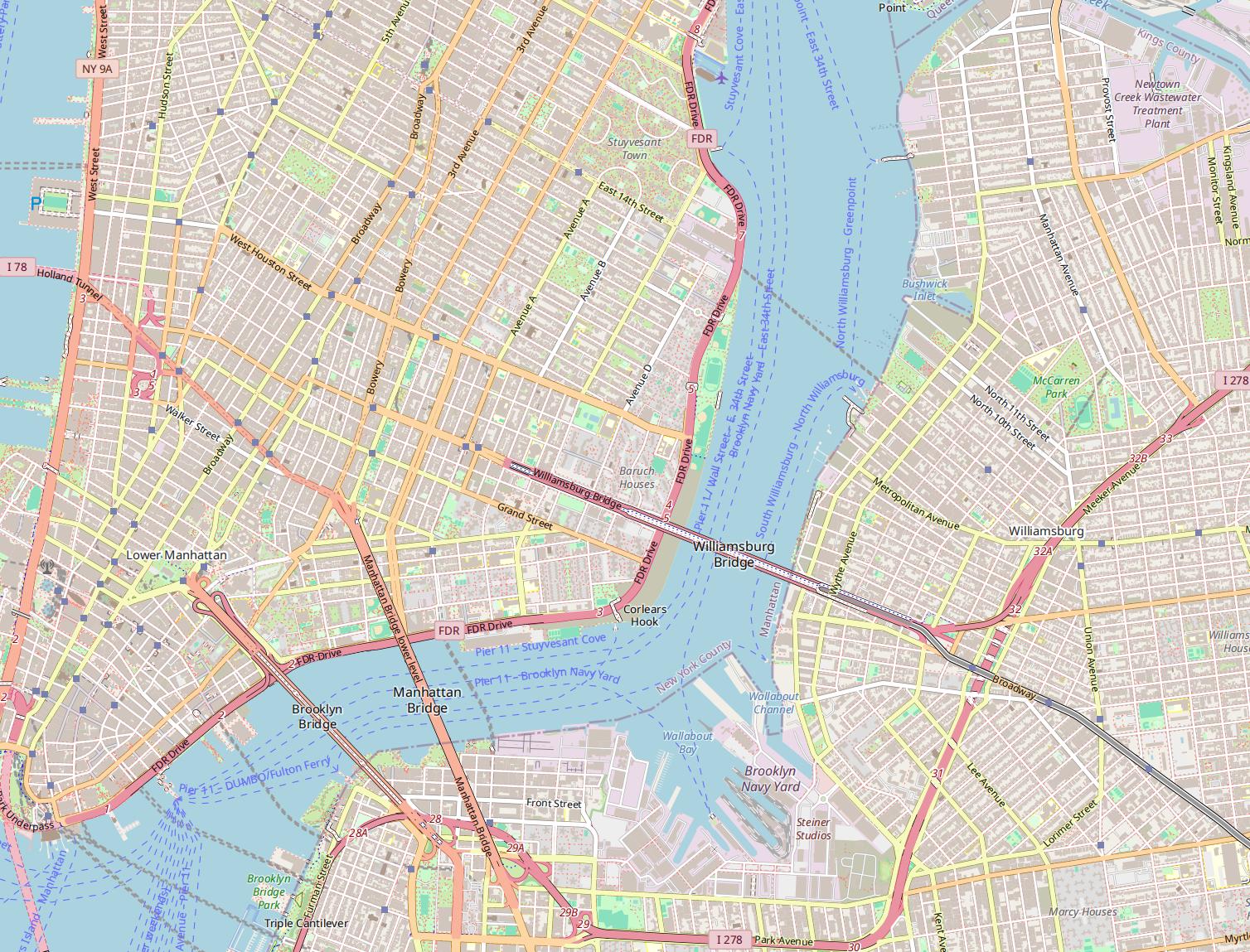} 
        \caption{New York}
        \label{fig:map_newyork}
    \end{subfigure}
    \caption{Maps for Beijing, London and New York.}
   \label{fig:map}
\end{figure*}

\begin{table}[h]
  \centering
  \setlength{\tabcolsep}{2pt} %
  \footnotesize %
  \renewcommand\arraystretch{1.1} %
  \caption{The raw data of the selected region in three cities.}
  \resizebox{0.48\textwidth}{!}{
  \begin{tabular}{lcccccc}
    \toprule
    \textbf{City} & \textbf{AoIs} & \textbf{PoIs} & \textbf{Roads} & \textbf{Trajectory} & \textbf{Street View Image} & \textbf{Satellite Image} \\
      \midrule
      \textbf{Beijing} & 4647 & 1882 & 2320 & 21015 & 28798 & 1533  \\
      \textbf{London}  & 13705 & 11715 & 1322 & 173268 & 3125 & 556  \\
      \textbf{New York} & 19541 & 11112 & 522 & 390934 & 24444 & 2738 \\
    \bottomrule
  \end{tabular}}
  
  \label{tab:orign-data}
\end{table}

\end{document}